\documentclass[10pt,journal]{IEEEtran}
\usepackage{amsmath,amsfonts,amssymb}
\usepackage{algorithmic}
\usepackage{array}
\usepackage[caption=false,font=normalsize,labelfont=sf,textfont=sf]{subfig}
\usepackage{textcomp}
\usepackage{stfloats}
\usepackage{url}
\usepackage{verbatim}
\usepackage{graphicx}
\def\BibTeX{{\rm B\kern-.05em{\sc i\kern-.025em b}\kern-.08em
    T\kern-.1667em\lower.7ex\hbox{E}\kern-.125emX}}
\usepackage{balance}
\usepackage{cite}
\usepackage{xcolor}
\usepackage{tabularx}
\usepackage{algorithm}
\usepackage{tikz}
\usepackage[binary-units]{siunitx}
\usepackage{bm}
\usepackage[normalem]{ulem}
\newcolumntype{Y}{>{\small\raggedright\arraybackslash}X}

\DeclareMathOperator*{\argmax}{arg\,max}

\usepackage{amsthm}
\usepackage{multirow}
\usepackage{float}
\usepackage{subfig}
\usepackage{svg}
\usepackage{caption}
\usepackage{subcaption}
\usepackage{pifont}
\usepackage{bbold}
\usepackage[colorlinks=true, allcolors=black]{hyperref}
\usepackage{setspace}

\title{
Federated Reinforcement Learning for Efficient Mobile Crowdsensing under Incomplete Information
}
\author{
Sumedh~J.~Dongare,~
Patrick~Weber,~
Andrea~Ortiz,~
Walid~Saad,~
Oliver~Hinz,~
Anja~Klein\\
\vspace*{-3mm}
\thanks{
Sumedh Dongare and Anja Klein are with the Communications Engineering Lab at the Technical University of Darmstadt, Germany.
Email:\{s.dongare, a.klein\}@nt.tu-darmstadt.de.\\
Patrick Weber and Oliver Hinz are with the Chair of Information Systems and Information Management at the Goethe University Frankfurt, Germany.
Email:\{weber, ohinz\}@wiwi.uni-frankfurt.de.\\
Andrea Ortiz is with the Institute of Telecommunications at the Vienna University of Technology, Austria.
Email:andrea.ortiz@tuwien.ac.at.\\
Walid Saad is with Wireless@VT, Bradley Department of Electrical and Computer Engineering, Virginia Tech, Alexandria, VA.
Email:walids@vt.edu\\
This work has been funded by the Deutsche Forschungsgemeinschaft (DFG, German Research Foundation) – Project-ID 210487104 - SFB 1053 MAKI. It also has been supported by the BMFTR project Open6GHub (Nr. 16KISK014).
The work of Andrea Ortiz is funded by the Vienna Science and Technology Fund (WWTF) [Grant ID: 10.47379/VRG23002].\\
Part of this work has been presented in \cite{Dongare_Globecom_2023}.
}
}

\theoremstyle{plain}

\newcommand{\mcite}[1]{\makebox[0.3cm]{\raisebox{0.15em}{\cite{#1}}}}

\begin{document}
\bstctlcite{IEEEexample:BSTcontrol}

\newlength\fwidth
\newlength\fheight
\renewcommand{\algorithmiccomment}[1]{\hfill{$\triangleright$\ #1}}
\newcommand{\rc}[1]{\textcolor{black}{#1}}

\maketitle
\begin{abstract}
Mobile crowdsensing (MCS) is a distributed sensing architecture that utilizes existing sensors on mobile units (MUs) to perform sensing tasks. A mobile crowdsensing platform (MCSP) publishes the sensing tasks and the MUs decide whether to participate in exchange for money. The MCS system is dynamic: the task requirements, the MUs' availability, and their available resources change over time. The MUs aim to find an efficient task participation strategy to maximize their income while the MCSP focuses on maximizing the number of completed tasks. As optimal strategies require perfect non-causal information about the MCS system, which is unavailable in realistic scenarios, the main challenge is to find an efficient task participation strategy for the MUs under incomplete information. To this end, a novel fully decentralized federated deep reinforcement learning algorithm, FDRL-PPO, is proposed. FDRL-PPO enables every MU to learn its own task participation strategy based on its experiences, available resources, and preferences, without relying on perfect non-causal information about the MCS system. To replenish their batteries, the MUs rely on energy harvesting. As a result, their available energy varies over time, leading to varying availability and fragmented learning experiences. To mitigate these challenges, the proposed approach leverages federated learning, enabling MUs to collaboratively improve their models without sharing private raw data like their own experiences. By exchanging only learned models, MUs collectively compensate for individual limitations, and find more scalable, robust, and efficient task participation strategies. Comprehensive evaluations on both synthetic and real-world datasets show that FDRL-PPO consistently outperforms benchmark algorithms in terms of task completion ratio, fairness in task completion, energy consumption, and number of conflicting proposals.
\end{abstract}

\vspace*{-3mm}
\section{Introduction}
\label{sec:Introduction}
In recent years, a novel sensing architecture called mobile crowdsensing (MCS) has emerged in the field of distributed sensing.
An MCS uses sensors already installed on smart devices (e.g. smartphones, wearables, and smart vehicles) to perform sensing tasks \cite{Ganti2011MCS_intro}.
Compared to the traditional wireless sensor networks (WSNs), MCS offers advantages including better coverage due to the mobility of mobile units (MUs), lower infrastructural costs and larger availability of MUs in a given area.
Consequently, MCS has become a topic of interest in the research community \cite{MCS_intro_Dai_21, iot_mcs_Jian_2015} and industry.
Many applications such as traffic monitoring \cite{Foursquare, komoot}, environmental monitoring \cite{Dinh_environmental_monitoring_2022}, spectrum sensing \cite{Spectrum_sensing_MCS_2021}, and mHealth \cite{Pryss2018mHealth} use MCS for distributed sensing.

An MCS system commonly consists of three entities: the data requesters, an MCS platform (MCSP), and the MUs.
The data requesters need sensing data from a geographical area and communicate this request to the MCSP.
The request includes different requirements of the sensing result (size in bits, deadline etc.) along with a predefined monetary budget for the sensing data.
The data requesters define this budget for each request based on the strictness of its requirements.
The MCSP reformulates the request to create a sensing task and broadcasts it together with its requirements to the MUs associated to the MCSP.
The MUs decide whether to indicate their willingness to perform an available task by means of a task proposal, or not.
Every MU's task proposal contains the task it wants to perform along with the desired payment which the MU expects as a reward.
MUs calculate their own desired payment based on their estimated task efforts without knowing the task budget.
As the MUs are typically battery operated, they require a charging mechanism to replenish their batteries.
In this work, we assume energy harvesting (EH) MUs for a sustainable MCS architecture \cite{Dongare_Globecom_2023, EH-WSNs_sandhu_2021}.

The MCSP responds to every proposal request with an accept or a reject signal.
If assigned, each MU performs the task independently.
In an ideal case, the MCSP would need only one MU to perform the sensing task.
However, to ensure task completion in a dynamic and uncertain MCS system, the MCSP may select multiple MUs within the task budget for redundancy.
Due to the limited task budget, the MCSP assigns each task to the cheapest MUs willing to perform it.
For every published task, the difference between the total number of proposing MUs and the number of selected MUs is the number of \textit{collisions}.
The collisions result in a degradation of performance because the rejected MUs have no task to perform in that time step and their sensing resources remain unused.
MUs which successfully complete the task by fulfilling its requirements receive their desired payment as reward.

The MUs and the MCSP need perfect non-causal information about the MCS system, i.e., complete information about the system dynamics for the current as well as the future time steps to make optimal task participation and assignment decisions, respectively.
On the one hand, the MUs need information about current and future tasks, i.e., task descriptions, the requirements, and the efforts required to perform them.
Moreover, since multiple MUs compete against each other to get assigned to the sensing task, every MU needs information about other MUs' participation strategies to optimize its own participation decisions and maximize its earning.
On the other hand, the MCSP requires information about MUs' desired payments for the current and future tasks in order to assign them to reliable MUs which will successfully complete them.
In real MCS systems, this perfect non-causal information is unavailable to both, the MCSP and the MUs.
This fact leads to information asymmetries, i.e.,
the MUs and the MCSP have different and incomplete information about the MCS system \cite{akerlof1970market, bergh2019information, sterz2022multi, Simon2025Bargaining}, thus addressing the task participation problem under incomplete information is of paramount importance.

\vspace*{-4mm}
\subsection{Related Works} \label{subsec:related_work}
\begin{table*}[ht]
\centering
\footnotesize
\caption{Comparison of related works in task allocation for MCS}
\label{tab:relatedWork}
\begin{tabular}{|>{\raggedright}m{4.8cm}|>{\centering\arraybackslash}m{0.3cm}|>{\centering\arraybackslash}m{0.3cm}|>{\centering\arraybackslash}m{0.3cm}|>{\centering\arraybackslash}m{0.3cm}|>{\centering\arraybackslash}m{0.3cm}|>{\centering\arraybackslash}m{0.3cm}|>{\centering\arraybackslash}m{0.3cm}|>{\centering\arraybackslash}m{0.3cm}|>{\centering\arraybackslash}m{0.3cm}|>{\centering\arraybackslash}m{0.3cm}|>{\centering\arraybackslash}m{0.3cm}|>{\centering\arraybackslash}m{0.3cm}|>{\centering\arraybackslash}m{0.3cm}|>{\centering\arraybackslash}m{0.3cm}|>{\centering\arraybackslash}m{0.3cm}|}
\hline
\textbf{Aspect} & \mcite{Dongare_Globecom_2023} & \mcite{Dongare_EHMCS_2022} & \mcite{C1:own:Dongare2024b} & \mcite{Bernd_WSaad_OSL_2024} & \mcite{OPAT_Huang_2022} & \mcite{Decentralized_TA_MARL_Xu_2023} & \mcite{Personalized_task_oriented_Wang_2021} & \mcite{MAB_TaskSelection_Sima_2022} & \mcite{Distributed_task_selection_Cheung_2021} & \mcite{Task_allocation_Time_Constraint_XinLi_2021} & \mcite{Sparse_MCS_MARL_Chunyu_2024} & \mcite{Mobility_prediction_MCS_fuzzy_Zhang_2023} & \mcite{UAV_assisted_MCS_Gao_2023} & \mcite{Bernd_ICC_2022} & \multicolumn{1}{m{0.55cm}|}{\textbf{This work}} \\ \hline

Only causal information required & \ding{51} & \ding{51} & \ding{51} & \ding{51} & - & \ding{51} & - & \ding{51} & - & - & \ding{51} & \ding{51} & \ding{51} & - & \ding{51} \\

Consideration of incomplete information & - & - & \ding{51} & \ding{51} & - & \ding{51} & \ding{51} & \ding{51} & \ding{51} & - & \ding{51} & - & - & - & \ding{51} \\

Consideration of MU preferences & \ding{51} & - & \ding{51} & \ding{51} & - & - & \ding{51} & \ding{51} & \ding{51} & - & - & - & - & \ding{51} & \ding{51} \\

Scalable approach & - & - & \ding{51} & \ding{51} & - & \ding{51} & \ding{51} & - & - & - & \ding{51} & - & - & \ding{51} & \ding{51} \\

Decentralized approach & \ding{51} & - & \ding{51} & \ding{51} & - & \ding{51} & \ding{51} & \ding{51} & \ding{51} & \ding{51} & \ding{51} & \ding{51} & \ding{51} & \ding{51} & \ding{51} \\

Robust to drop-outs and join-ins& \ding{51} & - & \ding{51} & \ding{51} & - & - & \ding{51} & - & \ding{51} & - & - & - & - & \ding{51} & \ding{51} \\

Energy, budget \& fairness constraints & \ding{51} & \ding{51} & - & - & \ding{51} & - & - & \ding{51} & \ding{51} & - & - & - & \ding{51} & - & \ding{51} \\ \hline

\end{tabular}
\end{table*}

In the prior art, many works assume the availability of perfect non-causal information at the MCSP regarding the complete MCS system when optimizing the task allocation decision \cite{OPAT_Huang_2022, Unified_JSCC_XLi_2023, Personalized_task_oriented_Wang_2021, Task_allocation_Time_Constraint_XinLi_2021}.
However, it is unrealistic to assume availability of such information about all MUs and all tasks.
\rc{In \cite{Unified_JSCC_XLi_2023}, the authors aim to jointly optimize the MU selection, bandwidth allocation, and sensing and transmission decisions using a unified approach that assumes complete information about the MCS system.}
The optimal centralized task allocation strategies often ignore the MU preferences and do not scale as the complexity of the problem grows.

To achieve scalability, previous works proposed decentralized game theoretic solutions that consider user preferences for the task proposal and task allocation problems \cite{Bernd_ICC_2022, Bernd_WSaad_OSL_2024, C1:own:Dongare2024b, OPAT_Huang_2022, Distributed_task_selection_Cheung_2021}.
Typically, these works assumed a snapshot based scenario in which stable matching solutions are obtained in every time step, independently of the previous ones.
To achieve these stable solutions, the MCSP requires private MU information, such as their task preferences, locations, and trajectories.
Many decentralized matching game solutions allow computationally feasible solutions.
\rc{For instance, in [16],[17], the authors consider a decentralized learning-guided matching approach to optimize task proposals and task assignments with incomplete information in an MCS scenario.
However, these works assume that the task proposal or assignment decisions have no consequences on future time steps.}

To overcome the requirement of perfect non-causal information, the works in \cite{Xu_Task_allocation_DRL_2023,Distributed_task_selection_Cheung_2021, C1:own:Dongare2024b, Bernd_WSaad_OSL_2024,Dongare_EHMCS_2022,MAB_TaskSelection_Sima_2022,Task_allocation_Time_Constraint_XinLi_2021, Mobility_prediction_MCS_fuzzy_Zhang_2023, Sparse_MCS_MARL_Chunyu_2024} employ learning solutions to improve the decisions made over time.
In \cite{Xu_Task_allocation_DRL_2023,Dongare_EHMCS_2022,MAB_TaskSelection_Sima_2022,Task_allocation_Time_Constraint_XinLi_2021, Mobility_prediction_MCS_fuzzy_Zhang_2023, Sparse_MCS_MARL_Chunyu_2024}, the authors rely on a simplifying assumption that periodic information exchange between the MUs and the MCSP improves their available information is considered.
The authors in \cite{Task_allocation_Time_Constraint_XinLi_2021,Sparse_MCS_MARL_Chunyu_2024, Mobility_prediction_MCS_fuzzy_Zhang_2023,UAV_assisted_MCS_Gao_2023, Dongare_EHMCS_2022} use centralized machine learning solutions to improve task allocation strategies of the MCSP.
However, centralized strategies do not scale and the learning performance deteriorates with an increasing number of available tasks, the number of MUs, and the considered length of the time horizon.
To overcome this issue, many works \cite{Dongare_Globecom_2023, MAB_TaskSelection_Sima_2022, C1:own:Dongare2024b, Bernd_WSaad_OSL_2024} consider decentralized learning strategies to improve task participation strategies of the MUs.
\rc{However, these works make overly simplified assumptions about the scenario, e.g., they assume that all MUs can participate in performing all tasks \cite{Dongare_Globecom_2023}, thus ignoring the fact that task can have different regions of interests or assume that the MUs have infinite battery capacity \cite{C1:own:Dongare2024b, Bernd_WSaad_OSL_2024}, or consider that the tasks have no deadlines \cite{MAB_TaskSelection_Sima_2022}.}
With the exception of our prior work \cite{Dongare_Globecom_2023}, fairness in task execution is not addressed in the reference schemes.
This is critical to ensure that MUs do not exclusively undertake less challenging tasks while maximizing the task completion ratio.
With decentralized learning solutions, the MU drop-outs and join-ins largely impact the performance of MCS which are unexplored in the literature.
In Table \ref{tab:relatedWork}, a comparison of the related works with respect to our contributions to this research area is provided.
In a nutshell, despite of addressing many important challenges, the design of an algorithm that is decentralized, learning-based, does not require sharing of private information, and moreover, which is robust to MU drop-outs and new join-ins under energy, budget and fairness constraints is still an open research topic.

\vspace*{-3mm}
\subsection{Contributions}
The main contribution of this work is the study of the task participation problem in a practical MCS system under incomplete information from a fully decentralized perspective.
Every MU aims to maximize its individual goal, i.e., its overall payment from the MCSP within a limited time horizon and under time, energy and budget constraints.
Such aim aligns with the global goal of maximizing the number of completed tasks in the system.
To achieve this, the MUs learn when to propose to a task and when to conserve energy, allowing other capable MUs to perform the tasks, based on their own experiences.
Note, however, that these experiences are usually limited because the MUs' finite battery restricts the number of tasks they can perform.
Furthermore, in case a new MU joins the MCS system, it starts from incomplete information and has to catch up with the other MUs which are more experienced in making task proposals.

To overcome these challenges, we propose an enhanced version of our previously developed federated deep reinforcement learning (FDRL-PPO) algorithm based on a multi-agent reinforcement learning framework.
\rc{FDRL-PPO enables every MU to learn its own efficient task participation strategy over time using its own experiences and the learned policies of other MUs, without disclosing any sensitive data \cite{Moudoud_FL_security_2024, houda_FL_security_2024, Nie_FLSecurity_2025}.
This is achieved using federation of policies by all agents instead of sharing privacy sensitive experiences such as their location, battery statuses, or channel information.}
Improving the original algorithm from \cite{Dongare_Globecom_2023}, we enable the MUs to independently choose how much they want to rely on their own experiences as compared to the experiences of the other MUs.
This approach is robust and adapts to MU drop-outs as the newly connected MUs benefit from the experiences of the older MUs and do not have to learn from scratch.

\rc{This journal is an extension of our previous conference paper [1].
    The main extensions, which constitute the contributions of this journal, are as follows:
    \begin{itemize}
        \item 
        We consider a realistic system in which in every time step,
        (a) the MCSP publishes multiple sensing tasks of various types following a task arrival rate, mimicking commercial MCS applications.
        (b) there are sensing tasks with predefined regions of interest (RoI) and sensing tasks without any RoI.
        Whether the task has RoI or not is decided by the DR.
        Based on the RoI of the task, the availability of the MUs varies over time. 
        (c) we model the joint allocation of communication and computing resources at every MU for task execution.
        This means, the MUs do not simply collect and send the sensing result back to the MCSP, but they process it before sending, which adds to their task execution efforts.
        \item For the considered MCS system, we formulate the task participation problem as a decentralized Markov game from MUs' perspective under incomplete information.
        \item We propose a novel, enhanced, and fully decentralized FDRL-PPO algorithm, which is implemented at every MU, and which does not rely on the availability of perfect non-causal information.
        On the contrary, federation allows MUs to share their learned policies with other MUs to develop better task participation policies while maintaining the competition to maximize their individual goals.
        \item We extensively validate the performance of our proposed approach using synthetically generated data as well as real-world data from a commercial MCS application.
        We evaluate task completion, collisions, fairness in task execution, and energy consumption as performance metrics.
        Simulation results demonstrate the superiority of our proposed FDRL-PPO algorithm across different scenarios.
        These scenarios analyze the impact of scalability, task load, task budgets, and MU join-ins and drop-outs on the performance. 
    \end{itemize}
}

The rest of the paper is organized as follows:
In Section \ref{sec:system_model}, the system model is introduced followed by the centralized reference problem formulation and decentralized reformulation of the same problem in Section \ref{sec:problem_formulation}.
Section \ref{sec:FDRL} introduces the proposed FDRL-PPO algorithm.
In subsections \ref{sec:simulation_results} and \ref{sec:dataset_evaluation}, we conduct the numerical evaluation based on the synthetic data set and the real-world data-set, respectively.
Section \ref{sec:conclusion} concludes the paper.

\vspace*{-3mm}
\section{System Model}
\label{sec:system_model}
We consider an MCS system which consists of an MCSP that sequentially publishes sensing tasks, and a set $\mathcal{K}$ of $K$ MUs with EH capabilities.
We divide time into $T$ discrete time steps with indices $t \in \{0, 1, \ldots, T-1\}$ of duration $\tau^{\text{int}}$ each.
In each time step $t$, the MCSP publishes a set $\mathcal{O}_t = \{O_{n,t}\}$ of tasks where $n \in \{0, 1,\ldots,N-1\}$ represents the task index.
These tasks can be, for example, sensing temperature, monitoring noise level, taking a picture, or taking a survey.

\begin{figure}[t]
\centering
\includegraphics[width=0.45\textwidth]{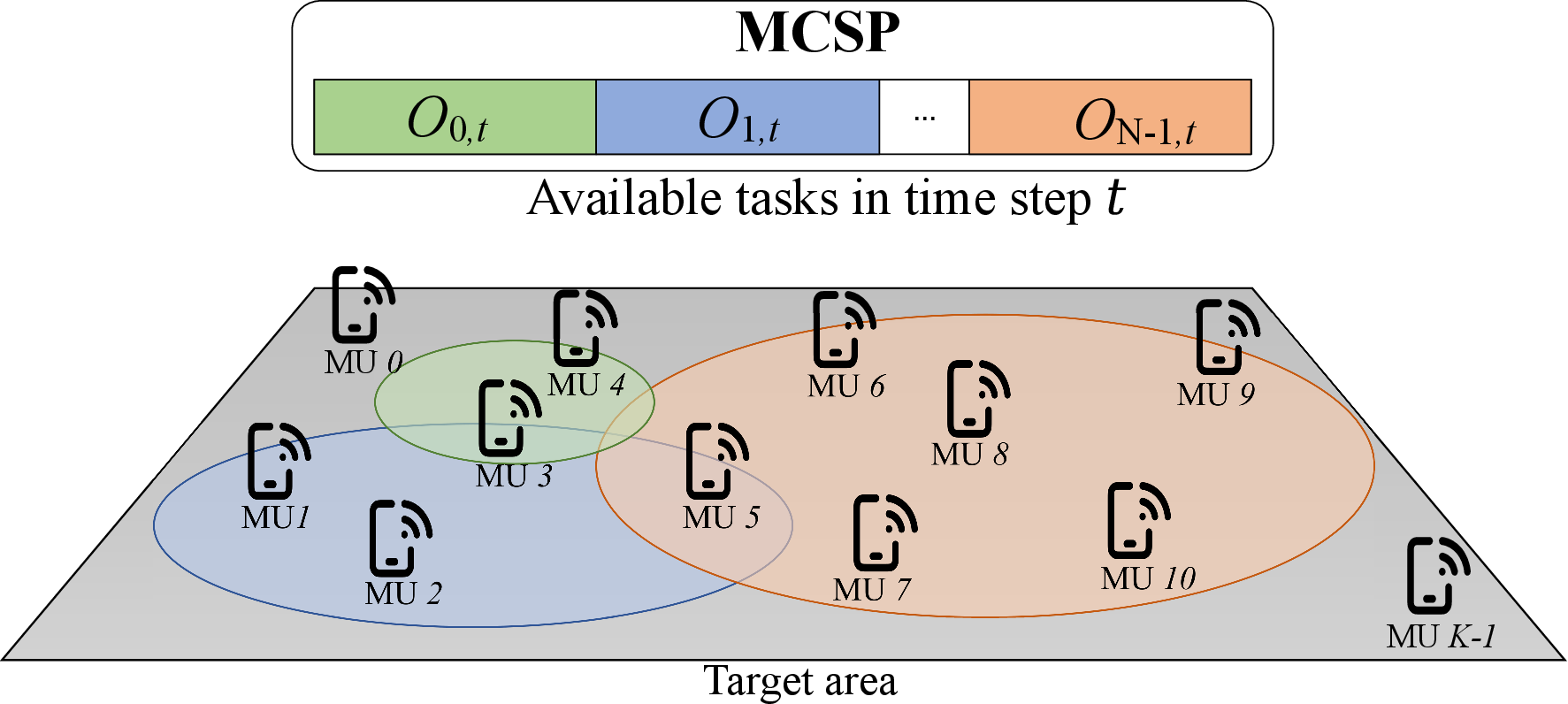}
\caption{A sample MCS system model}
\label{fig:system_model}
\vspace*{-1mm}
\end{figure}

Each sensing task in our system has a specific RoI.
The MU has to be within the RoI to be able to perform the sensing task and generate a valid sensing result.
The RoI may vary for different sensing tasks, e.g., for some sensing tasks the area can be very small and for some sensing tasks, it can be as large as the target area, allowing all of the associated MUs to propose and perform the sensing task, if accepted by the MCSP.
Fig. \ref{fig:system_model} shows a snapshot of the system in time step $t$.
The MCSP publishes $N$ sensing tasks with their specific RoIs.

\vspace*{-3mm}
\subsection{Mobile Crowdsensing Platform}
At the beginning of every time step $t$, the MCSP publishes a set $\mathcal{O}_t$ of sensing tasks.
Task $O_{n,t} \in \mathcal{O}_t$ is characterized by its requirements: $\langle M_{n,t}, \tau^\text{dl}_{n,t}, A_{n,t}^{i,j} \rangle$, where $M_{n,t}$ is the task size in bits and $\tau^\text{dl}_{n,t}$ is the deadline.
The RoI of task $O_{n,t}$ in time step $t$ is given by its coordinates $i$ and $j$ in the target area, and denoted by $A_{n,t}^{i,j}$.
Each task $O_{n,t}$ is classified into a task type $c\in\{0, 1, \ldots, C-1\}$, e.g., sensing temperature or taking a picture.
Each task type is characterized by the average size of the sensing result.
All tasks of the same type have similar sizes but may vary in terms of deadline and RoI.

Based on the task requirements, the MCSP defines a weight $V_{n,t}$, which represents the difficulty in executing task $O_{n,t}$ from the MCSP's perspective 
\begin{equation}
\label{eq:v_t_calculation}
    V_{n,t} = \mathbb{1}_{\{A_{n,t}^{i,j} = L_{k,t}^{i,j}\}}(\xi M'_{n,t} + \omega \tau^\text{dl'}_{n,t}),
\end{equation}
where $M'_{n,t}$ is the normalized sensing result size given by $\frac{M_{n,t}}{M_\mathrm{max}}$, with $M_\mathrm{max}$ being the size of the largest sensing task, and $\tau^\text{dl'}_{n,t}$ defined as the complement of normalized task deadline given by $1-(\frac{\tau^\text{dl}_{n,t}}{\tau^\text{int}})$.
The indicator function has value $1$ if the location of the MU $L_{k,t}^{i,j}$ is within the task RoI denoted by $A^{i,j}_{n,t}$, otherwise 0.
Note that the $i$ and $j$ are the coordinates of these locations.
The variables $\xi, \omega \in [0, 1]$ are weighting factors whose values trade the importance of the task size and  deadline in deciding the overall task difficulty, with $\xi+\omega=1$.
For a task with larger sensing result size $M_{n,t}$, the MUs would need to spend more resources to collect and process the sensing data successfully.
Therefore $V_{n,t}$ increases as $M_{n,t}$ increases.
In contrast to this, as the task deadline $\tau^\text{dl}_{n,t}$ gets shorter, the task execution becomes more difficult, and hence, $V_{n,t}$ increases.
Based on the difficulty $V_{n,t}$ of task $O_{n,t}$, the data requesters set a task budget $Z^{\mathrm{task}}_{n,t}=\eta V_{n,t}$, where $\eta$ is the budget coefficient.
This is the maximum amount of monetary units the MCSP can spend for the execution of task $O_{n,t}$.

At the beginning of every time step $t$, the MUs convey their willingness to perform task $O_{n,t}$ along with their desired payments $G_{k,n,t}$.
Without loss of generality, we assume that one MU can perform only one sensing task in one time step and therefore proposes to only one task in a time step.
The MCSP collects all the proposals and selects the cheapest MUs for task execution within the task budget $Z^{\mathrm{task}}_{n,t}$.
Multiple MUs can be selected to perform the task depending on the task budget.
The MCSP does so for redundancy and to have more confidence in the results sent by the MUs and account for measurement errors of the MUs.
It suffices to the MCSP if at least one of the selected MUs performs the task successfully to consider it completed.
The platform's decision $x_{k,n,t} \in \{0,1\}$ is sent back to the MUs.
$x_{k,n,t}=1$ denotes that MU $k$ has been selected by the MCSP to perform the task $O_{n,t}$, and $x_{k,n,t}=0$ indicates that it has not been selected.
All the task assignment decisions are stored in the matrix $\mathbf{X}=(\mathbf{x}_{0,0}, \ldots, \mathbf{x}_{N-1,T-1})$, where $\mathbf{x}_{n,t}=(x_{0,n,t},\ldots,x_{K-1,n,t})^\mathrm{T}$.
Note that the rejected MU does not know, whether and how many other MUs' proposals are accepted by the MCSP.
Every MU knows about the rejection or acceptance their own proposal only.
The assigned MUs then perform the sensing task and generate sensing data.
They process this sensed data to obtain the sensing result and transmit this result to the MCSP over a wireless channel.
The MUs can still fail the task at this point due to uncertain quality of the wireless channel.
The MCSP pays $G_{k,n,t}$ to MU $k$ for successfully completing the task $O_{n,t}$.

\vspace*{-5mm}
\subsection{Mobile Units}
\begin{table}[t]
\centering
\caption{Notations}
\scriptsize
\label{Table:Notation}
\begin{tabular}{|l|l|}
\hline
\textbf{Description} & \textbf{Notation} \\ \hline
Time step index & $t$ \\
Duration of one time step & $\tau^{\text{int}}$ \\
Total number of time steps in given time horizon & $T$ \\
\hline
\textbf{Task Characteristics:} & \\
Region of Interest (RoI) of task $O_{n,t}$ & $A_{n,t}^{i, j}$ \\
Type of task $O_{n,t}$ & $c\in\mathcal{C}$ \\
Size of task $O_{n,t}$ before processing & $\Bar{M}_{n,t}$ \\
Size of task $O_{n,t}$ after processing & $M_{n,t}$ \\
Task index & $n$ \\
Total number of tasks published in time step $t$ & $N$ \\
$n$th task published in time step $t$ & $O_{n,t}$ \\
Set of available tasks in time step $t$ & $\mathcal{O}_t$ \\
Task difficulty weight of task $O_{n,t}$ & $V_{n,t}$ \\
Task assignment matrix & $\mathbf{X}$ \\
Task budget of task $O_{n,t}$ & $Z^{\mathrm{task}}_{n,t}$ \\
Task processing complexity & $\delta_{n,t}$ \\
Deadline of task $O_{n,t}$ & $\tau^{\mathrm{dl}}_{n,t}$ \\
\hline

\textbf{MU Characteristics:} & \\
Battery status of MU $k$ in time step $t$ & $b_{k,t}$ \\
Battery capacity of each MU & $B_{\mathrm{max}}$ \\
Amount of harvested energy by MU $k$ in time step $t$ & $E^{\mathrm{harv}}_{k,t}$ \\
Energy required for proposing to task $O_{n,t}$ & $E^{\mathrm{p}}_{k,n,t}$ \\
Sensing energy of MU $k$ for task $O_{n,t}$ & $E^{\mathrm{s}}_{k,n,t}$ \\
Transmission energy of MU $k$ for task $O_{n,t}$ & $E^{\mathrm{tx}}_{k,n,t}$ \\
Computing energy of MU $k$ for task $O_{n,t}$ & $E^{\mathrm{comp}}_{k,n,t}$ \\
Total execution energy of MU $k$ for task $O_{n,t}$  & $E^{\mathrm{exec}}_{k,n,t}$ \\
Desired payment of MU $k$ for task $O_{n,t}$ & $G_{k,n,t}$ \\
Channel coefficient for MU $k$ for task $O_{n,t}$ & $h_{k,t}$ \\
Set of available MUs & $\mathcal{K}$ \\
MU index & $k$ \\
Total number of available MUs & $K$ \\
Location of MU $k$ in time step $t$ & $L_{k,t}^{i,j}$ \\
Channel bandwidth for MU $k$ & $W$ \\
Sensing time of MU $k$ for task $O_{n,t}$ & $\tau^{\mathrm{s}}_{k,t}$ \\
Computing time of MU $k$ for task $O_{n,t}$ & $\tau^{\mathrm{comp}}_{k,t}$ \\
Transmission time of MU $k$ for task $O_{n,t}$ & $\tau^{\mathrm{tx}}_{k,t}$ \\
Total execution time of MU $k$ for task $O_{n,t}$ & $\tau^{\mathrm{exec}}_{k,t}$ \\
\hline

\textbf{MU Decisions:} & \\
Task participation matrix & $\mathbf{Y}$ \\
Computing resource allocation matrix & $\mathbf{F^\mathrm{comp}}$ \\
Transmit power allocation matrix & $\mathbf{P^{\mathrm{tx}}}$ \\
\hline
\end{tabular}
\end{table}
Each MU $k$ makes three consecutive decisions in each time step $t$: task participation decision, selection of the amount of computing resources for processing of the sensing data, and selection of transmit power to send the processed sensing result back to the MCSP.
The location of MU $k$ in time step $t$ is $L_{k,t}^{i,j}$ where $i$ and $j$ represent its coordinates.

We define $y_{k,n,t} \in \{0, 1\}$ as the task participation decision where $y_{k,n,t}=1$ indicates that MU $k$ is willing to perform the task $O_{n,t}$ from the set $\mathcal{O}_t$ of available tasks, and $y_{k,n,t}=0$ indicates otherwise.
If assigned by the MCSP, i.e. $y_{k,n,t}=x_{k,n,t}=1$, the total effort in terms of execution energy $E^\text{exec}_{k,n,t}$ put in by MU $k$ for task $O_{n,t}$ will be
\vspace{-0.5em}
\begin{equation}
    E^\text{exec}_{k,n,t}=E^\text{p}_{k,n,t}+E^\text{s}_{k,n,t}+E^\mathrm{comp}_{k,n,t}+E^\text{tx}_{k,n,t},
\end{equation}
where $E^\text{p}_{k,n,t}$ is the energy spent by MU $k$ to send the proposal for task $O_{n,t}$ over the wireless channel in $t$, $E^\text{s}_{k,n,t}$ is the energy required for sensing, $E^\text{comp}_{k,n,t}$ is the energy required for processing the sensing data, and $E^\text{tx}_{k,n,t}$ is the transmission energy required to send the sensing result back to the MCSP.
Similarly, the total execution time required for MU $k$ in $t$ is given by 
\vspace{-0.5em}
\begin{equation}
    \tau^\text{exec}_{k,n,t}=\tau^\text{s}_{k,n,t}+\tau^\mathrm{comp}_{k,n,t}+\tau^\text{tx}_{k,n,t},
\end{equation}
where $\tau^\text{s}_{k,n,t}$ is the sensing time required to collect sensing data, $\tau^\text{comp}_{k,n,t}$ is the time required for processing this sensing data, and $\tau^\text{tx}_{k,n,t}$ is the transmission time required to send the sensing result back to the MCSP.
Since $\tau^\text{exec}_{k,n,t}$ starts after the task assignment from the MCSP, the time required to propose to a task is not considered.

The sensing energy is given by $E^\text{s}_{k,n,t} = \tau^\text{s}_{k,n,t}p^\text{s}_{k,n,t}$, where $p^\text{s}_{k,n,t}$ is the required power for sensing.
The sensing time $\tau^\text{s}_k$ is a random variable with mean value $\Bar{\tau}^\text{s}_k$ for each MU $k$.
The actual sensing time $\tau^\text{s}_{k,n,t}$ for each MU $k$ to generate a sensing result for task $O_{n,t}$ in time step $t$ depends on the characteristics of the MU's sensor and its conditions in time step $t$.
Similarly, the sensing power $p^\text{s}$ is a random variable with mean value of $\Bar{p^\text{s}}$.
The actual sensing power $p^\text{s}_{k,n,t}$ for MU $k$ to sense for task $O_{n,t}$ in time step $t$ depends on its task type $c$ and the sensor specific conditions of MU $k$ in time step $t$. 

The computing resources $f_{k,n,t}^\mathrm{comp}$ in CPU cycles/s are used for processing of raw sensing data $\Bar{M}_{k,n,t}$ such that it can be converted into the required size $M_{n,t}$ of the sensing result, such that $\Bar{M}_{k,n,t}>M_{n,t}$.
$f_{k,n,t}^\mathrm{comp}$ can take any value in the range 
$(0, f_\textrm{max}^\mathrm{comp}]$ where 
$f_\textrm{max}^\mathrm{comp}\in\mathbb{R}$ is the 
maximum computing cycles that can be assigned per second \cite{OPAT_Huang_2022,C1:own:Dongare2024b, Bernd_WSaad_OSL_2024}.
By deciding the computing resource $f_{k,n,t}^\mathrm{comp}$, the expenditure of computing energy is calculated as $E_{k,n,t}^\mathrm{comp}=\tau^\mathrm{comp}_{k,n,t}p^\mathrm{comp}_{k,n,t}$, where $p^\mathrm{comp}_{k,n,t}$ is the computing power required for the processing.
The computing time $\tau^\mathrm{comp}_{k,n,t}$ of task $O_{n,t}$ is
\vspace*{-2mm}
\begin{equation}\label{eq:computing_time}
    \tau^\mathrm{comp}_{k,n,t} = \frac{\Bar{M}_{k,n,t}\delta_{n,t}}{f_{k,n,t}^\mathrm{comp}},
\end{equation}
where $\delta_{n,t}$ is the task processing complexity in CPU cycles/bit.

The transmit power $p^\text{tx}_{k,n,t}$ is used by MU $k$ for the transmission of the task proposal, and if selected for task execution, then for transmitting the sensing result of task $O_{n,t}$.
$p^\text{tx}_{k,n,t}\in\mathbb{R}$ can take any value in the range $[0, p^\text{tx}_\text{max}]$ where $p^\text{tx}_\text{max}$ is the maximum transmit power.
With this selected transmission power $p^\mathrm{tx}_{k,n,t}$, MU $k$ controls the expenditure of transmission energy $E^\text{tx}_{k,n,t} = \tau^\text{tx}_{k,n,t}p^\text{tx}_{k,n,t}$, where $\tau^\text{tx}_{k,n,t}$ is the transmission time required to communicate $M_{n,t}$ bits from MU $k$ back to the MCSP for task $O_{n,t}$ defined by
\vspace{-0.5em}
\begin{equation}\label{eq:shannon_eq}
\tau^{\text{tx}}_{k,n,t} = \frac{M_{n,t}}{W\log_2\left(1 +\frac{p^{\text{tx}}_{k,n,t}|h_{k,t}|^2}{\sigma^2}\right)},
\end{equation}
where $W$ is the channel bandwidth in MHz, $h_{k,t}$ is the channel coefficient for the link between the MCSP and MU $k$ in time step $t$ and $\sigma^2$ is the noise power.

The payment request $G_{k,n,t}$ is proportional to the effort $E^\text{exec}_{k,n,t}$ made by each interested MU $k$ willing to perform the task $O_{n,t}$ such that, $G_{k,n,t}=\kappa E^\text{exec}_{k,n,t}$, where $\kappa$ is a factor in monetary units per Joule.
Note that MUs have no incentive to falsely ask for a larger payment since this reduces their chances of getting accepted by the MCSP.
Thus, competition between the MUs compels them to behave honestly since they wish to maximize long term rewards.

Each MU $k$ harvests an amount of energy $E^\text{harv}_{k,t}$ in Joules in every time step $t$.
This energy is stored in a battery with capacity $B_\text{max}$ without any losses.
The MUs update their battery status $b_{k,t}$ at the end of time step $t$ as
\begin{equation}\label{eq:battery_update}
    b_{k,t} = \min\left\{B_\text{max}, b_{k,t-1} - E^\text{exec}_{k,n,t} + E^\text{harv}_{k,t}\right\}.
\end{equation}
To ensure energy causality, each MU $k$ may only use the amount of energy available in the battery.
The notations are summarized in Table \ref{Table:Notation}.

\vspace*{-3mm}
\section{Problem formulation}
\label{sec:problem_formulation}
\subsection{Centralized task allocation problem with complete information as a reference}
We will first formulate the centralized task allocation problem considered as an optimization problem from the MCSP's perspective.
This is to obtain a performance upper bound in terms of the number of completed tasks over a finite time horizon.
To solve this problem, we assume that the MCSP has perfect non-causal information about the battery statuses, the amounts of harvested energy, the communication channel conditions, the available computing resources of all MUs, and the descriptions of all tasks.
Note that this solution is used as a reference scheme to compare the performance upper bound.

The MCSP may now exploit this available information to 
centrally and optimally decide which MU to choose for each task as well as determine the computing resources and transmit power this MU should select to successfully complete the task.
Since the MCSP has complete information, it needs to assign the task to only one MU in order to successfully complete the task reliably and efficiently.
Due to the limited battery capacity and poor channel conditions, the MCSP may also choose not to perform a certain task and save the MUs' energy for better, more suitable future tasks.
The goal of task allocation is to maximize the average number of completed tasks weighted by the respective factor $V_{n,t}$.
In this case, the MUs do not have any preference since the MCSP optimally allocates a suitable task $O_{n,t}$ for each MU $k$ by exploiting the perfect non-causal information.
This problem is NP-hard and grows exponentially as the time horizon, number of available tasks, and number of MUs increase \cite{OPAT_Huang_2022, Distributed_task_selection_Cheung_2021, Dongare_EHMCS_2022, Dongare_Globecom_2023}.
The three decisions pertaining to task allocation, transmit power selection, and computing resource allocation are stored in the matrices $\mathbf{Y}$, $\mathbf{P^\text{tx}}$, and $\mathbf{F^\mathrm{comp}}$ respectively.
These matrices follow a structure $\mathbf{Y}=(\mathbf{y_{0,0},y_{0,1},\ldots,y_{N-1,T-1}})$, where $\mathbf{y_{n,t}}=(y_{0,n,t},y_{1,n,t},\ldots,y_{K-1,n,t})^\text{T}$.
For the computing resources, $\mathbf{F^\mathrm{comp}}=(\mathbf{f^\mathrm{comp}_{0,0}}, \mathbf{f^\mathrm{comp}_{0,1}}, \ldots, \mathbf{f^\mathrm{comp}_{K-1,T-1}})$, where $\mathbf{f^\mathrm{comp}_{k,t}}=(f^\text{comp}_{k,0,t}, f^\text{comp}_{k,0,t}, \ldots, f^\text{comp}_{k,N-1,t})^\text{T}$.
Similarly, $\mathbf{P^\text{tx}}=(\mathbf{p^\text{tx}_{0,0},p^\text{tx}_{0,1},\ldots,p^\text{tx}_{N-1,T-1}})$, where $\mathbf{p^\text{tx}_{n,t}}=(p^\text{tx}_{0,n,t},p^\text{tx}_{1,n,t},\ldots,p^\text{tx}_{K-1,n,t})^\text{T}$.
The task $O_{n,t}$ can be assigned to potentially all of the $K$ available MUs.
\vspace{-0.5em}
\begin{equation}\label{eq:requiredMUConstraint1}
    \sum_{k=0}^{K-1} y_{k,n,t} \leq K,\quad \forall n,\forall t. 
\end{equation}
Note that the MCSP has to also take into account the task budget which is explained in the later in the section.
Each MU $k$ can be assigned to only one task $O_{n,t}$ in time step $t$
\vspace{-0.5em}
\begin{equation}\label{eq:requiredMUConstraint2}
    \sum_{n=0}^{N-1} y_{k,n,t} \leq 1,\quad \forall k,\forall t. 
\end{equation}
Since the tasks have a specific RoI, the MUs must recognize if they are within this RoI to be eligible to participate in those task executions.
Note that for some tasks, the RoI can be as large as the target area.
In such cases, all MUs are potentially capable of participating in the task.
A deadline constraint for each task $O_{n,t}$ in time step $t$ has to be fulfilled by the allocated MU $k$, i.e.,
\vspace{-0.5em}
\begin{equation}\label{eq:deadline_constraint}
    \sum_{k=0}^{K-1}\tau^\text{exec}_{k,n,t}y_{k,n,t} \leq \tau^\text{dl}_{n,t}, \forall n,\forall t.
\end{equation}
To satisfy this constraint, the MCSP optimally evaluates the computing resources $f^\mathrm{comp}_{k,n,t}$ and transmit power $p^\text{tx}_{k,n,t}$ for MU $k$.
The transmit power $p^\text{tx}_{k,n,t}$ is constrained by
\begin{equation}\label{eq:power_constraint}
    0 \leq p^\text{tx}_{k,n,t} \leq p^\text{tx}_{\mathrm{max}} \quad \forall k,\forall n,\forall t.
\end{equation}
The constraint on computing resource $f_{k,n,t}^\mathrm{comp}$ \cite{OPAT_Huang_2022, TMahn2021} is
\begin{equation}\label{eq:computing_resource_constraint}
    0 < f_{k,n,t}^\mathrm{comp} \leq f_{\mathrm{max}}^\mathrm{comp} \quad \forall k,\forall n,\forall t.
\end{equation}
The budget constraint restricts the MCSP to allocate the task only to the MUs that fit within the task budget $Z^{\text{task}}_{n,t}$ and is given by
\vspace*{-3mm}
\begin{equation}\label{eq:Budget_constraint}
    \sum_{k=0}^{K-1}G_{k,n,t}y_{k,n,t} \leq Z^{\text{task}}_{n,t}, \quad \forall n,\forall t.
\end{equation}
The energy causality constraint
\begin{equation}
    \sum_{j=0}^{J-1} E^{\text{harv}}_{k,j} - \sum_{j=1}^{J}\sum_{n=0}^{N-1}E^{\text{exec}}_{k,n,j} y_{k,n,t} \geq 0, \forall k, J = 1, \ldots, T ,\label{eq:Energy_Causality}
\end{equation}
guarantees that MU $k$ does not spend more than available energy.
Moreover, the MUs cannot spend $E^{\text{harv}}_{k,t}$ in the same time step $t$, but only in the later time steps.
\begin{equation}
    \sum_{j=0}^{J - 1} E^{\text{harv}}_{k,j}  - \sum_{j=1}^{J}\sum_{n=0}^{N-1} E^{\text{exec}}_{k,n,j} y_{k,n,t} \leq B_{\text{max}}, \forall k, \forall J \label{eq:Battery_Overflow}
\end{equation}
indicates battery overflow constraint which ensures that the maximum value of energy that can be stored in the battery is $B_\mathrm{max}$.
The optimization problem from the MCSP's perspective to maximize the average weighted sum of completed tasks is
\begin{equation} \label{eq:optimization_problem}
\begin{split}
\argmax_{\{y_{k,n,t}, f^\mathrm{comp}_{k,n,t}, p_{k,n,t}^{\text{tx}}\}} & \quad \sum_{t=0}^{T-1}\sum_{n=0}^{N-1} V_{n,t} \sum_{k=0}^{K-1}y_{k,n,t} \\
\text{subject to} & \quad (\ref{eq:requiredMUConstraint1}) - (\ref{eq:Battery_Overflow}).
\end{split}
\end{equation}
These constraints are interdependent and non-convex.
To fairly handle requirements of different tasks, we maximize the number of completed tasks weighted by the factor $V_{n,t}$.

\vspace*{-3mm}
\subsection{Problem Reformulation as Decentralized Markov Game}
\label{subsec:MarkovGame}
We formulate the centralized optimization problem in Section \ref{sec:problem_formulation} from the MCSP's perspective.
To optimally solve this problem, the MCSP needs the perfect non-causal information.
However, in realistic scenarios, neither MCSP nor MUs have perfect non-causal information about the MCS system to make optimal decisions.
They have to make decisions under incomplete information.
To overcome the limitations of centralized solutions, we reformulate the optimization problem in (\ref{eq:optimization_problem}) as a decentralized task participation problem from the perspective of the MUs.
The MUs are incentivized to earn more money through task participation to fulfill the goal of maximizing the average weighted number of completed tasks.
Every MU makes an independent decision on its own task participation based on the current information it has about its own battery status, channel conditions, available computing resources, and about the requirements of available tasks.
Due to the decentralized nature of the problem and as the considered system exhibits Markov property, we model it as a Markov Game with MUs as the rational decision makers which aim to maximize their individual rewards.

Each MU $k$ makes decisions about its own task participation in every time step $t$, denoted by $y_{k,n,t}$, its computing resource allocation $f^\mathrm{comp}_{k,n,t}$, and the transmit power selection $p_{k,n,t}^\mathrm{tx}$.
Such problems in which multiple agents make individual decisions based on the information available to them, can be formulated as a Markov Game (MG).
An MG is characterized by a tuple $\langle \mathcal{K}, \mathcal{S}_k, \mathcal{A}_k, \mathcal{R}, \gamma \rangle$, where $\mathcal{K}$ is the set of players in the game, which in our case is the set of available MUs.
The sets $\mathcal{S}_0, \mathcal{S}_1,\ldots,\mathcal{S}_{K-1}$ consist of individual states which can be observed by the respective MU $k\in\mathcal{K}$.
A state $S_{k,t}\in\mathcal{S}_k$ is the available information at MU $k$ which it uses to select action $A_{k,t}\in\mathcal{A}_k$ in time step $t$.
Every MU can perform certain actions collected in the set $\mathcal{A}_k$ to maximize its income.
The possible rewards are collected in the set $\mathcal{R}\rightarrow\mathcal{S}\times\mathcal{A}_1\times\mathcal{A}_2\ldots\times\mathcal{A}_K$ for each possible state-action pair for every MU.

In the scope of this work, we define the state $S_{k,t}\in\mathcal{S}$ observed by MU $k$ in time step $t$ as $S_{k,t}= \langle b_{k,t}, \Bar{h}_{k,t}, L_{k,t}^{i,j}, M_{n,t}, c_{n,t}, \tau^\mathrm{dl}_{n,t}\rangle$ for all tasks with indices $n\in{0,1,\ldots,N-1}$. $\Bar{h}_{k,t}$ is the average channel coefficient calculated from the past channel coefficients monitored by MU $k$ in time step $t$.
State $S_{k,t}$ contains the information available at MU $k$ which it utilizes to take a suitable action $A_{k,t}$, with $A_{k,t}=\langle y_{k,n,t}, f^\mathrm{comp}_{k,n,t}, p^\mathrm{tx}_{k,n,t} \rangle$.
Accounting for constraints in (\ref{eq:requiredMUConstraint1}-\ref{eq:requiredMUConstraint2}) and (\ref{eq:power_constraint}-\ref{eq:computing_resource_constraint}), the action space of every MU is constrained within feasible values.
For the action $A_{k,t}$, every MU $k$ receives a reward $R_{k,t}\in\mathcal{R}$ if the deadline constraint (\ref{eq:deadline_constraint}) and budget constraint in (\ref{eq:Budget_constraint}) are successfully followed.
In our scenario, $R_{k,t}=G_{k,n,t}$ if MU $k$ successfully completes task $O_{n,t}$.
The energy causality constraint (\ref{eq:Energy_Causality}) and battery overflow constraint (\ref{eq:Battery_Overflow}) are naturally followed since in sequential execution, the MU cannot spend more energy than it already has and cannot store more energy than its battery capacity $B_\mathrm{max}$.

Each MU $k$ aims to maximize the long-term discounted reward $R=\sum_{t=0}^\infty\gamma^tR_{k,t}$, where $\gamma\in[0,1]$ is the discount factor.
Every MU $k$ achieves this goal by learning a task participation policy $\pi^k$ which maps the action $A_{k,t}$ on to the state $S_{k,t}$ by using an artificial neural network.
\vspace*{-3mm}
\section{Reinforcement Learning (RL) Solution}
\label{sec:FDRL}

\subsection{PPO-based task participation strategy}
\label{subsec:PPO}
To address the continuous state and action space of each MU, we implement the deep reinforcement learning algorithm proximal policy optimization (PPO) \cite{schulman2017proximal} at each MU $k$ and train it based on their independent experiences.
Specifically, we use PPO with gradient clipping, known as PPO-clip since it ensures that the policies are not changed drastically in the training iterations resulting in a stable, efficient, and generalizable task participation policy at the MU.
Without loss of generality, we use the terms PPO and PPO-clip interchangeably hereinafter.
PPO is an actor-critic policy gradient method consisting of two networks, the actor (or policy) network, and the critic (or value) network.
The actor decides which action $A_{k,t}$ should be taken in a given state $S_{k,t}$ based on MU $k$'s policy $\pi^k(A_{k,t}|S_{k,t}; \theta^k)$.
The critic informs the actor how good or bad this action $A_{k,t}$was and how to improve it by computing the value function $V^{\pi^k}(S_{k,t};\phi^k)$.

To train our proposed approach, we consider multiple training episodes.
A training episode $i$ consists of $T$ time steps in which each MU observes its own states, takes its individual actions, and observes the individual rewards. The observed states, selected actions, and rewards within one training episode $i$ are termed the trajectory $\mathcal{D}^k_i$.
At the beginning of the first training episode $i=1$, each MU $k$ initializes its policy parameters $\theta^k$ and value function parameters $\phi^k$.
These parameters are updated in each subsequent training episode using the observed trajectories $\mathcal{D}^k_i$.
Specifically, at the end of each training episode, the policy network (i.e. the actor) updates the parameters $\theta^k$ in the direction that maximizes the average long term rewards via stochastic gradient ascent algorithm.
Similarly, the value network (i.e. the critic) updates $\phi^k$ by minimizing the loss function which represents the gap between the expected long term rewards and the actual rewards using stochastic gradient descent algorithm. 
The same procedure is repeated until convergence is achieved.

\subsection{Federated Deep RL using PPO}
In the previous section, we explained the PPO algorithm.
In the considered MCS system, each MU should ideally learn to which tasks it should propose and to which tasks it should not propose and let other (potentially better suitable) MUs participate.
This is an example of IA reduction with the help of screening \cite{stiglitz1977monopoly}.
However, with no communication between the MUs, this is difficult to learn.
The considered MCS system is mixture of competitive and cooperative environment since the MUs potentially compete against each other to get tasks assigned to them by the MCSP.
However, they have to cooperate with each other to learn about the tasks faster by exchanging information about the tasks \cite{jiang2016share, akerlof1970market}.
By cooperating with each other, MUs can collectively learn to maximize their own payments and also maximize the average weighted sum of completed tasks in a finite time horizon.
This cooperation is achieved by sharing their learnt models, i.e., $\theta_k$ and $\phi_k$.
This information exchange also helps the newly associated MUs because they can take advantage of the already learned models of other MUs.
Note that by sharing the learning parameters, the MU's private and sensitive data is preserved and not shared.
Our proposed approach enables this sharing by exploiting a federated learning algorithm to distributively train the MUs using PPO.
\begin{algorithm}[t]
    \caption{FDRL-PPO}\label{alg:FDRL_PPO}
    \begin{algorithmic}[1]
        \begin{scriptsize}
        \STATE \textbf{Initialization:}
        \STATE Initialize a global model with weights $\Omega(0)$ at the MCS platform.
        \STATE Each MU $k$ initializes a local model $\Omega_k(0)$ and set it with global model weights such that $\Omega_k(0)=\Omega(0),\forall k\in\mathcal{K}$.
        \FOR{each round $r = 0,\ldots,r_\mathrm{max}-1$}
            \FOR{each MU $k = 0, 1, \ldots, K-1$}
                \STATE Download global model $\Omega(r)$ from the platform.
                \STATE Set local model $\Omega_k(r)=\chi_k\Omega_k(r-1)+(1-\chi_k)\Omega(r-1)$
                \STATE Train the model locally using weights $\Omega_k(r)$.
                \COMMENT Section \ref{sec:FDRL}
                \STATE Upload weights after training to the MCS platform.
            \ENDFOR
            \STATE \textbf{At the MCSP:}
            \STATE Collect all weight updates from all MUs.
            \STATE Compute federated averaging to obtain $\Omega(r+1)$
            \algorithmiccomment{Eq.(\ref{eq:fedavg})}
        \ENDFOR
        \end{scriptsize}
    \end{algorithmic}
\end{algorithm}
We present this in Algorithm \ref{alg:FDRL_PPO}.

In FDRL-PPO training phase, an aggregator node (in our case the MCSP) initializes a global model $\Omega(r=0)$, consisting of the policy and critic network parameters $\theta$ and $\phi$, respectively (Line 2).
The aggregator node is not necessarily the MCSP, but we assume so for simplicity.
In practice, it can be a separate entity.
Moreover, the MCSP only aggregates the global model without accessing model details or private MU information.
Secure updates can be ensured through privacy-preserving mechanisms like encryption, but this is beyond the scope of this work.
We divide each training episode $i$ into $r_\mathrm{max}$ federation rounds indexed by $r$, such that each federation round $r$ is formed by $T_r$ time steps.
Every MU $k$ initializes its own local model $\Omega_k(0)$ with weights identical to the globally initialized model $\Omega(0)$ (Line 3).
In every round $r$, every MU $k$ downloads the last updated global model $\Omega(r-1)$ (Line 6).
Using global model, every MU $k$ updates its local model (Line 7)
\vspace*{-0.5em}
\begin{equation}
    \Omega_k(r)=\chi_k\Omega_k(r-1)+(1-\chi_k)\Omega(r-1).
\end{equation}
Here, $\chi_k=\frac{R_{k,r}}{\sum_k R_{k,r}}$ is a reward based update factor which every MU $k$ maintains to strategically decide how much should the global model influence its local model.
Note that every MU uses its local model for decision making and the global model is generated for improving the learning speed and generalization capability of MUs.
The update factor is higher when the MU is new to the MCS system and wants to learn faster.
However, this factor can change over time such that MUs rely more on their own experiences and still have generalizable models to develop policies.
Then every MU $k$ trains its local model through its own decisions and experiences accumulated over the rest of the federation round (Line 8).
At the end of round $r$, each MU $k$ transmits this trained local model $\Omega_k(r)$ to the MCSP (Line 9).
Once the MCSP receives all the local models (Line 12), it performs reward weighted federated averaging based on the rewards of each agent to generate the updated global model (Line 13)
\vspace{-0.5em}
\begin{equation}
\label{eq:fedavg}
    \Omega(r+1) = \sum_{k=0}^{K-1}\frac{R_{k,r}}{\sum_k R_{k,r}}\Omega_k(r).
\end{equation}
This way, the aggregated model $\Omega(r+1)$ is closer to the best performing MU.
The MUs download the updated global model $\Omega(r+1)$ and retrain this model based on their local data.

\vspace*{-4mm}
\subsection{Computational complexity of FDRL-PPO}
\label{subsec:complexity}
The computational complexity of the proposed FDRL-PPO algorithm does not grow with the number of MUs due to its distributed architecture, which is beneficial in real-world applications.
However, the complexity of the aggregator, which in our work is the MCSP, has complexity that grows linearly with the number of MUs $K$.
We assume that each MU's forward and backward pass through its policy network costs $\Psi$ floating‐point operations (FLOPs), which depends on the dimensions of the network.
Every MU collects $T_r$ experiences in one federation round $r$ requiring $O\bigl(\Psi\,T_r)$ FLOPs at every MU~\cite{schulman2017proximal}.
In every federated averaging step, each MU can communicate its full policy network once per round to the MCSP~\cite{mcmahan2023communicationefficientlearningdeepnetworks}, making the complexity of the aggregation at the MCSP grow linearly with $K$.
At the MUs, over $r_\mathrm{max}$ federation rounds the total complexity becomes $O\bigl(r_\mathrm{max}\,T_r\,\Psi)$.

\vspace*{-4mm}
\subsection{Convergence and Stability of FDRL-PPO}
\label{subsec:convergence_stability}
In this section, we discuss the convergence and stability properties of the proposed FDRL-PPO.
As with many reinforcement learning algorithms, the convergence of FDRL-PPO to a global optimum cannot be guaranteed due to the infinitely large state and action spaces resulting in a infinite-dimensional policy space \cite{PPO_Convergence_and_optimality_Liu_2019, PPO_convergence_problem_2002, schulman2017proximal}.
\rc{While PPO provides a monotonic improvement guarantee on a surrogate objective under certain assumptions in the single-agent setting \cite{schulman2017proximal}, such guarantees do not directly extend to multi-agent environments, where the learning dynamics are inherently non-stationary due to multiple simultaneously learning agents.
Consequently, the stability of the solution to the formulated multi-agent game cannot be theoretically guaranteed.}

\rc{In the proposed FDRL-PPO algorithm, each MU seeks to improve its own expected return based on local observations.
However, the federated learning mechanism enables information sharing across MUs through a global model, which helps to reduce policy drift and mitigates non-stationarity in practice.
By partially aligning local model updates through federation, agents benefit from a more informative and consistent global model, leading to more stable learning dynamics compared to fully decentralized baselines.
The simulation results demonstrate that the proposed FDRL-PPO algorithm exhibits stable and robust learning behavior in realistic MCS scenarios indicating its strong practical potential.
}
\vspace*{-3mm}
\section{Simulation Results and Analysis}
\label{sec:numerical_evaluation}
\begin{table}
  \caption{Simulation parameters}
  \label{tab:evaluation_parameters}
	 \centering{
  \scriptsize
	 \def\arraystretch{1.2}
	\begin{tabular}{|l|l|}
	\hline
	\textbf{Parameter} & \textbf{Value}  \\
	\hline
	Total number of time steps $T$ & $1000$ time steps \\
	Duration of one time step $t=\tau^\mathrm{int}$ & $\SI{10}{\second}$\\
	Number of available MUs $K$ & [5, 50]\\
    Number of available tasks $N$ per $t$ & [5, 50] tasks \\
    Number of different task types & 10 \\
	MU distances to MCSP $[d_{\text{min}},d_{\text{max}}]$ & $[100,1000]\,\SI{}{\metre}$ \\
	Battery capacity $B_{\text{max}}$ & $\SI{8}{\watt\second}$ \\
	Max. harvested energy $E^\mathrm{harv}_{\text{max}}$ per $t$ & $10\%$ of $B_{\text{max}}$\\
	\hline
	Total Bandwidth $W$ per MU $k$ & $\SI{1}{\mega\hertz}$ \\
	Noise power $\sigma^2$ & $10^{-16}\,\SI{}{\watt}$ \\
	Transmit power $p^\mathrm{tx}_{\text{max}}$ of sensor $k$ \cite{TMahn2021} & $\SI{200}{\milli\watt}$ \\
	Channel gain $|h_{k,t}|^2$ & $\sim d^{-3}$\\
	\hline
    Size of sensed data $\Bar{M}_{n,t}$ \cite{OPAT_Huang_2022} & [20, 200] \SI{}{\mega\bit}\\
	Sensing task result size $M$ \cite{OPAT_Huang_2022} & $[1-8]$ \SI{}{\mega\bit} \\
	Task deadline $\tau^\mathrm{dl}$ & $[\frac{\tau^\mathrm{int}}{2},\tau^\mathrm{int}] \SI{}{\second}$\\
    Task processing complexity $F_{n,t}$ \cite{OPAT_Huang_2022} & [200, 300] $\frac{\text{CPU cycles}}{\text{bit}}$\\
    MU CPU clock frequency $f^\mathrm{comp}_{k}$ \cite{OPAT_Huang_2022} & [200, 400] \SI{}{\mega\hertz}\\
	\hline
	\end{tabular}}
\end{table}
In this section, we present the numerical evaluation of the performance of the proposed FDRL-PPO algorithm in comparison with the reference schemes.
We first analyze the performance using a synthetically generated dataset with controlled parameters to avoid biased results.
Then, we validate the performance using a real-world dataset.

\vspace*{-4mm}
\subsection{Evaluation using Synthetic Dataset}
\label{sec:simulation_results}
The simulation results are averaged over $I=100$ independent realizations.
In each realization, we consider $T=1000$ time steps of length $10$ seconds.
In each time step, we vary the number of available tasks in the set $\mathcal{O}_t=\{O_{n,t}\}$.
The MUs are in an area where the maximum distance between an MU and the MCSP is $1$ km.
The communication channel between MU $k$ and the MCSP is modelled as a Rayleigh fading channel with path loss exponent of three.
The MUs move freely in the area with a maximum average speed of $\SI{10}{\kilo\metre}/\text{h}$.
We set $\xi$ and $\omega$ from (\ref{eq:v_t_calculation}) to one to promote fair task completion.
Table \ref{tab:evaluation_parameters} lists our simulation parameters.

In the proposed FDRL-PPO framework, hyperparameter optimization is applied to ensure training stability and performance. The final configuration includes a learning rate of $l_r=5e^{-5}$, a training batch size of $100$ time steps, a mini-batch size of $64$. Each agent’s policy and value networks employ two hidden layers with $256$ neurons per layer and ReLU activations. The Adam optimizer with a decaying learning rate schedule is used to balance exploration and exploitation.

We consider following reference schemes for comparison.\\
\textbf{Optimal task allocation (OTA):} As described in Section \ref{sec:problem_formulation}, this is a centralized task allocation scheme which assumes perfect non-causal information about the amount of harvested energy, communication channel conditions, and the future tasks and their requirements.
Although unrealistic, this scheme provides the performance upper bound.\\
\textbf{Myopically optimal task participation (MOTP):} This approach assumes that each MU has perfect causal information about its communication channel coefficients.
Based on this, each MU $k$ makes an informed decision on task participation, computing resource allocation, and transmit power selection.
Thus, this scheme makes myopically optimal decisions for the current time step neglecting the future consequences.\\
\textbf{Random task participation scheme (RTPS):} A light-weight task participation scheme in which every MU $k$ randomly proposes to the available tasks, and randomly assigns resources for the task execution when selected by the MCSP.
It ensures fairness in task completion across the types as the MUs propose to all types with equal probability.
\
\textbf{Independent PPO (IPPO)~\cite{IPPO_deWitt_2020}:} A deep reinforcement learning scheme based on PPO where each MU $k$ improves its own task participation policy based on local available information.
\\
\rc{\textbf{Multi-agent PPO (MAPPO)~\cite{yu2022surprising}:} A deep reinforcement learning scheme based on centralized-training-decentralized-execution (CTDE) framework where one MU develops and trains its policy based on global state and reward information.
This information is acquired by sharing states and actions through explicit communication channels between the MUs. 
In the validation phase, the same trained policy is shared with all the MUs. Although it is unrealistic to implement in practical MCS applications, we use this algorithm as a benchmark.}

\rc{In all the reference schemes except for OTA and MAPPO, every MU is unaware of the task participation strategy of other MUs. 
In OTA, the MCSP centrally and optimally assigns the tasks to suitable MUs based on their capability. Whereas in MAPPO, a shared policy is used in validation phase.}

For numerical evaluation, we develop $4$ different scenarios.
\textbf{\rc{Scenario 1 (Baseline scenario):}} This is a baseline scenario in which the platform publishes five tasks per time step, for a time horizon of $100$ time steps.
Five MUs are available to propose and participate in the sensing task execution.

\begin{figure*}[t]
   \centering
    \subfloat[][\footnotesize{Avg. weighted completed tasks}]{
    \includegraphics[width=0.24\linewidth]{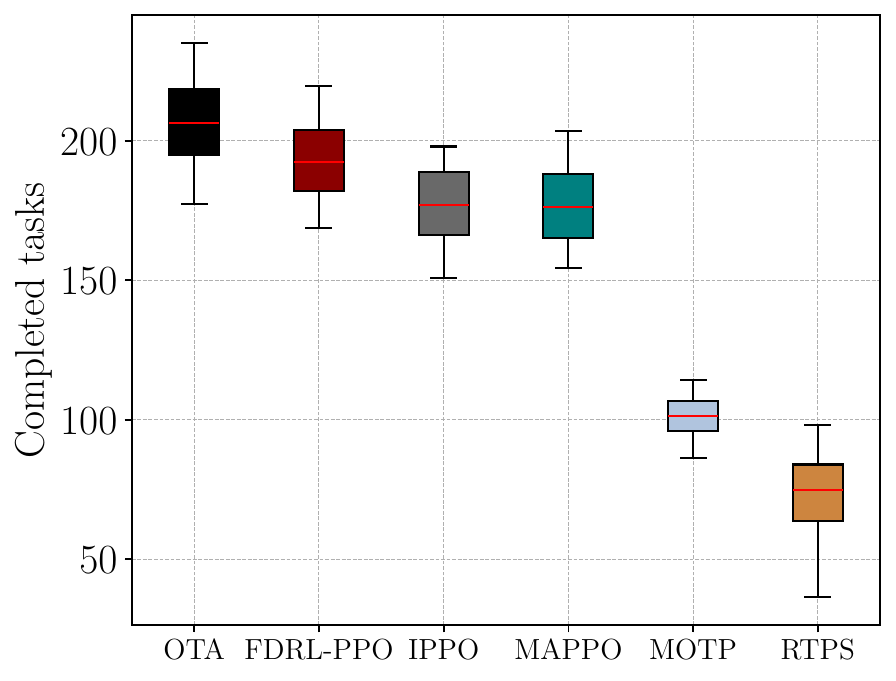}
    \label{fig:basline_performance_comparison}
    }
    \subfloat[][\footnotesize{Average collision ratio}]{
    \includegraphics[width=0.24\linewidth]{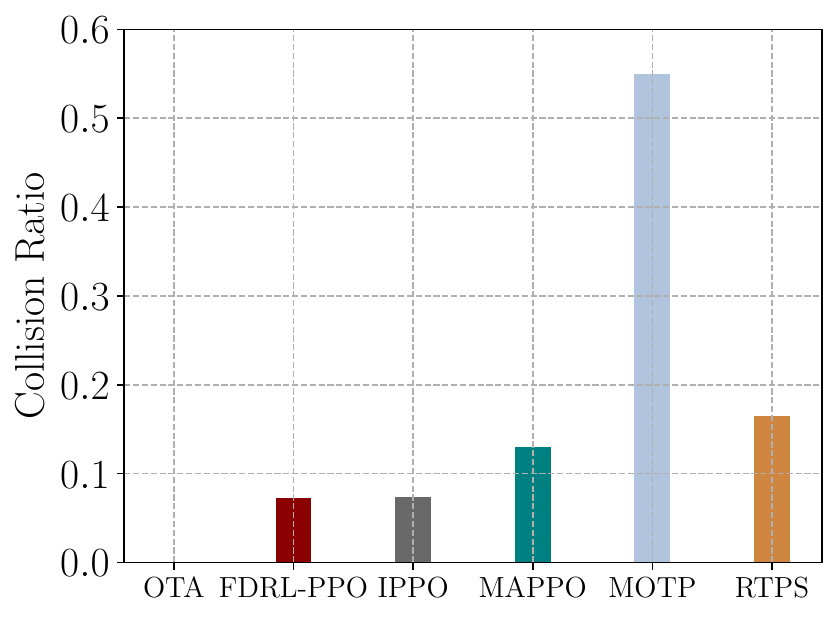}
    \label{fig:basline_collisionRatio}
    }
    \subfloat[][\footnotesize{Average completed tasks for each task type}]{
    \includegraphics[width=0.24\linewidth]{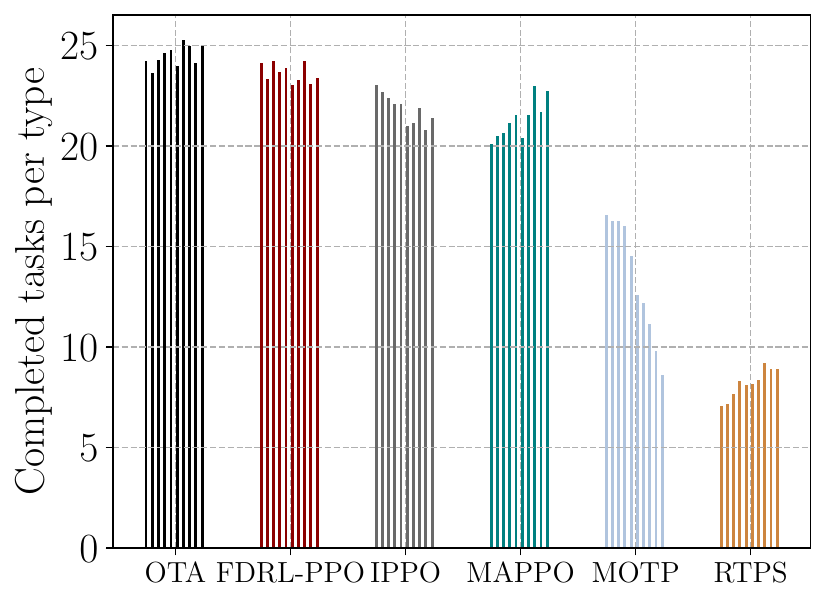}
    \label{fig:basline_completed_tasks_per_tasktype}        
    }
    \subfloat[][\footnotesize{Energy consumption}]{
    \includegraphics[width=0.24\linewidth]{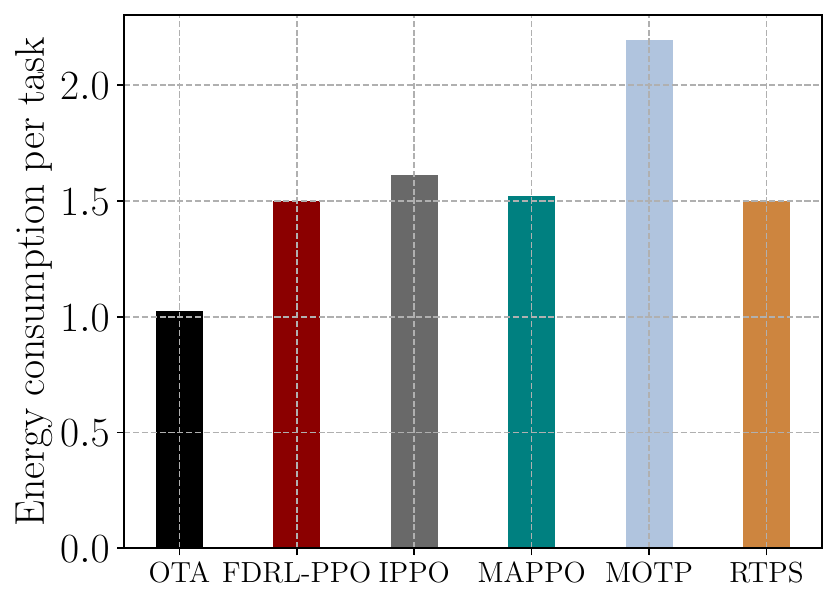}
    \label{fig:basline_energy_consumption}        
    }
    \vspace*{-2mm}
    \caption{Scenario 1: Comparison of different performance metrics for baseline scenario}
    \label{fig:baseline_scenario_synthetic_data}
    \vspace*{-4mm}
\end{figure*}

\rc{In Fig.~\ref{fig:baseline_scenario_synthetic_data}, we compare the performance of our FDRL-PPO algorithm with respect to the average number of weighted completed tasks, average collisions normalized by the number of available MUs, task completion based on the task types, and the energy consumption against the performance of the reference algorithms.
Particularly, for the comparison of the average number of weighted completed tasks, we use box-plots in which minimum and maximum values are depicted along with $5\%$ and $95\%$ confidence intervals across all the independent realizations.
The mean value is depicted with a red line within the box-plot.
In Fig.~\ref{fig:basline_performance_comparison}, we see that the performance of OTA is superior to all of the other schemes.
To achieve this, OTA uses the perfect non-causal information about the scenario.
In comparison to this, our proposed FDRL-PPO performs only $4\%$ below the OTA and outperforms IPPO by $8.4\%$, MAPPO by $8.3\%$, MOTP by $51\%$, and RTPS by $68.75\%$, without requiring the perfect non-causal information and additionally considering the preferences of the MUs.
This is because with FDRL-PPO, the MUs learn a task participation strategy which helps them to decide in which tasks they should participate, and in which they should not.
This is also evident in Fig.~\ref{fig:basline_collisionRatio}, where we compare the average collision ratio, i.e., the ratio of collisions per total number of task proposals in all of the algorithms including our proposed FDRL-PPO approach.
We see that since the OTA performs centralized task allocations, there are no collisions.
In contrast, our FDRL-PPO along with IPPO experience $8\%$ collisions per MU on average.
Due to its greedy nature, MOTP has the highest average collision ratio per MU.
Similarly, MAPPO has $12\%$ collisions on average.
This is because when MUs visit similar states, MAPPO forces the MUs to take similar actions, resulting in collisions.
Since the RTPS randomly participates in any of the available tasks, the average collision ratio is lower than the MOTP, however, still higher than our proposed FDRL-PPO approach.}

\rc{In Fig.~\ref{fig:basline_completed_tasks_per_tasktype}, we evaluate whether the task execution is fair with respect to the task types.
In the figure, each strategy displays $10$ bars representing each task type from easy to difficult with respect to the factor $V$.
Like OTA, FDRL-PPO performs all the tasks fairly.
In contrast, IPPO performs more easy tasks than difficult ones.
This is even more significant in the task execution of MOTP.
In contrast, MAPPO performs difficult tasks more than the easy ones.
This is because the difficult task types offer higher budget and thus, more MUs can get accepted to those task types.
Similarly the RTPS also performs more difficult tasks than easy ones.
For RTPS, this effect is a combined result of the overestimation of desired payments and the task budgets with respect to the task types.
For difficult tasks, the budget is higher and thus, MUs may get assigned to these tasks easily if they decide to participate.
Therefore, in RTPS, MUs get assigned to difficult tasks more than easy tasks, which reflects in the task execution behavior.}

\rc{The energy consumption of all algorithms normalized by the total number of completed tasks is evaluated in Fig. \ref{fig:basline_energy_consumption}.
Since OTA requires perfect non-causal information, it precisely knows how much energy is required to complete each task, and thus has the least energy expenditure.
The MOTP, being a greedy approach, spends high energy per task without consideration for the future time steps.
As MAPPO and RTPS perform difficult tasks more than the easy ones, their average energy expenditure over completed tasks is comparable to that of our proposed FDRL-PPO.
However, this does not mean that MAPPO and RTPS are more efficient than our FDRL-PPO.
On the contrary, FDRL-PPO maximizes the average weighted completed tasks while minimizing the collisions and the energy consumption and is fair to all task types in their execution.
FDRL-PPO outperforms IPPO and MOTP by $6.7\%$ and $46.7\%$, respectively and w.r.t. energy consumption.}

\begin{figure}[t]
   \centering
    \begin{minipage}[c]{\linewidth}
    \vspace{-2mm}
    \centering
    \includegraphics[width=\linewidth]{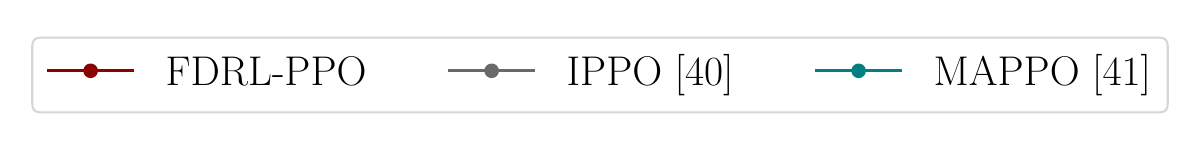}
    \vspace*{-6mm}
    \end{minipage}
    \begin{minipage}[c]{0.49\linewidth}
        \includegraphics[width=\linewidth]{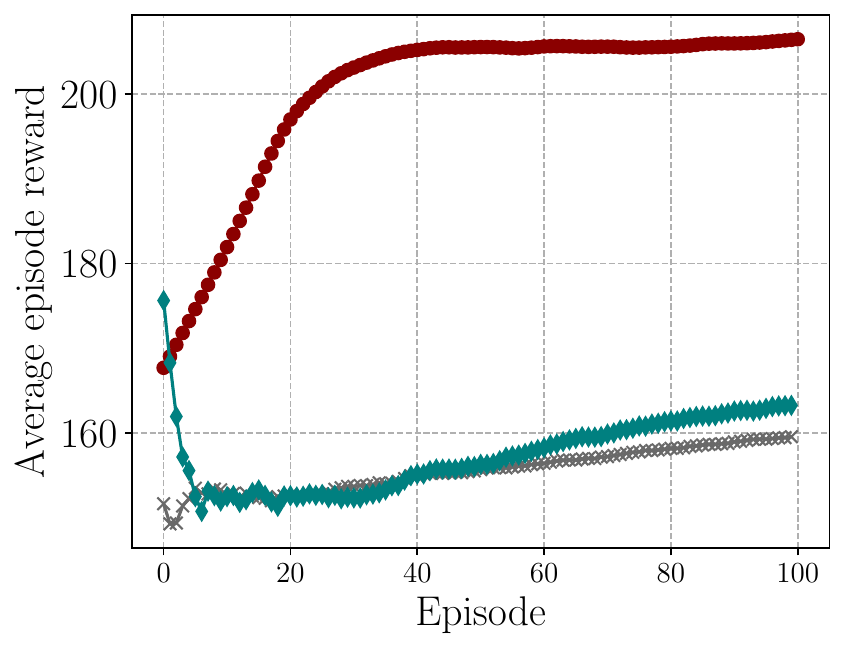}
        \vspace*{-7mm}
        \caption{\rc{Training performances of RL-based benchmarks}}
        \label{fig:training_progress}
    \end{minipage}
    \hfill
    \begin{minipage}[c]{0.48\linewidth}
        \includegraphics[width=\linewidth]{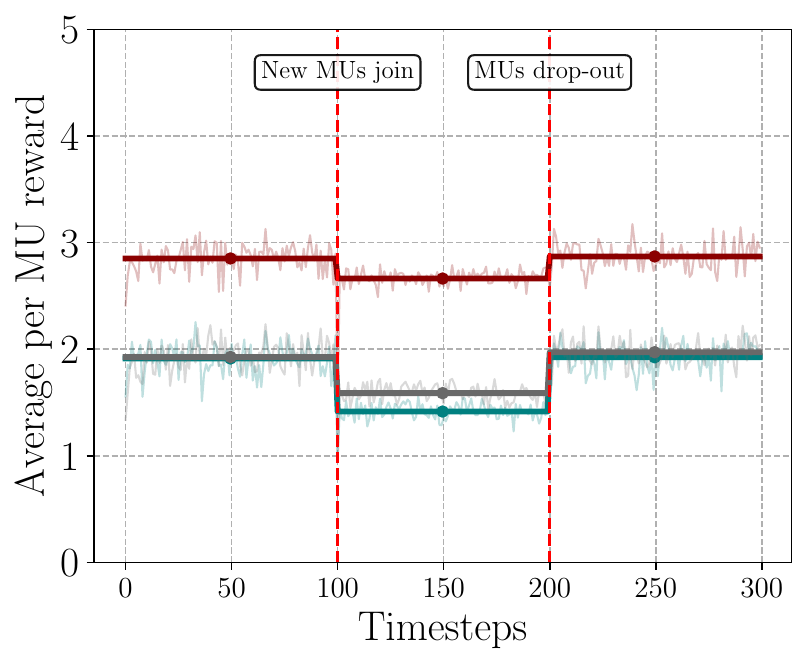}
        \vspace*{-7mm}
        \caption{\rc{MU-join-in and MU-drop-out analysis over time}}
        \label{fig:MU_join_in_drop_out_analysis}
    \end{minipage}
\end{figure}
\rc{
In Fig.~\ref{fig:training_progress}, we compare the training performance of the FDRL-PPO algorithm with the considered RL-based benchmarks.
We see that the FDRL-PPO algorithm achieves a stable performance after $40$ episodes.
In comparison, the IPPO and the MAPPO algorithms struggle to improve the learning policies over episodes.
Although theoretical stability guarantees cannot be provided for the FDRL-PPO algorithm, this training performance highlights the effectiveness of the federated learning in non-stationary multi-agent scenarios.
}

\rc{In Fig.~\ref{fig:MU_join_in_drop_out_analysis}, we analyze the effect of MU-join-ins and MU-drop-outs on the performance of our FDRL-PPO algorithm in comparison with the considered RL-based benchmarks.
For this evaluation, we consider a time horizon of $T=300$ time steps.
In the training phase, we assume that $K=5$ MUs were present. 
Later in the validation phase, we assume that the same five MUs are present for the first $100$ time steps.
After $t=100$, we assume that five new MUs join the MCS system. 
After $t=200$, randomly five MUs drop out of the MCS system. 
These events are marked with vertical dotted lines in Fig.~\ref{fig:MU_join_in_drop_out_analysis}.
We compare the average per MU reward over time for this analysis. 
We observe that overall, FDRL-PPO demonstrates a better and more robust performance in comparison with the benchmarks. 
For the time interval $t=[100, 200]$, we see a drop in the average reward per MU for all the algorithms. 
This is because in this interval, there are more MUs competing with each other and thus, the individual average reward of each MU reduces. 
However, for the proposed FDRL-PPO, the performance drop is only $3\%$ as compared to $15\%$ drop in IPPO and $30\%$ drop in MAPPO algorithms. 
This is because the FDRL-PPO algorithm enables the newly joined MUs to inherit the global model directly instead of learning from scratch. 
In comparison, the other algorithms exhibit poor performance since the newly joined MUs have untrained task participation policy.}

\rc{
While MAPPO is a strong multi-agent RL baseline, it typically entails centralized training and significant communication overhead to collect and utilize global information.
Under our distributed deployment assumptions, these requirements limit its practical relevance.
Additionally, MAPPO and RLIA yield comparable performance in our experiments, whereas FDRL-PPO consistently achieves the best results.
We therefore exclude MAPPO from subsequent plots to improve readability and avoid repetitive comparisons.}

\textbf{\rc{Scenario 2 (MU Scalability analysis):}} In this scenario, we analyze the effect of the number of MUs on the average weighted number of completed tasks, average collision ratio, and the number of completed tasks per task type.
In each time step, the MCSP publishes $15$ tasks of different types.
The scenario runs for $T=1000$ time steps.
The number of MUs are varied in the range $[10, 50]$ with a step size of $10$ MUs.
Since the computing time for such a large scenario is too high for OTA, we omit this algorithm from the analysis for the rest of the scenarios.

\begin{figure*}[t]
    \begin{minipage}[c]{\linewidth}
    \centering
    \includegraphics[width=0.6\linewidth]{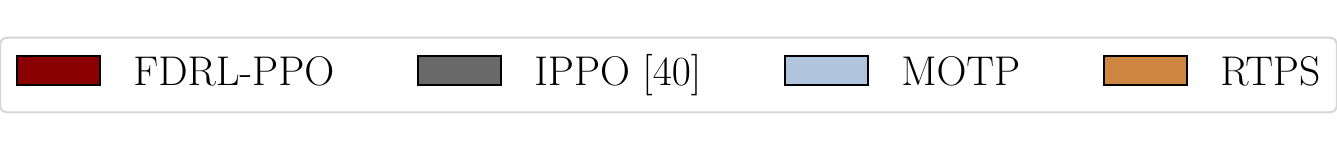}
    \vspace*{-6mm}
    \end{minipage}
   \centering
       \subfloat[\footnotesize{Average weighted completed tasks}]{
       \includegraphics[width=0.25\linewidth]{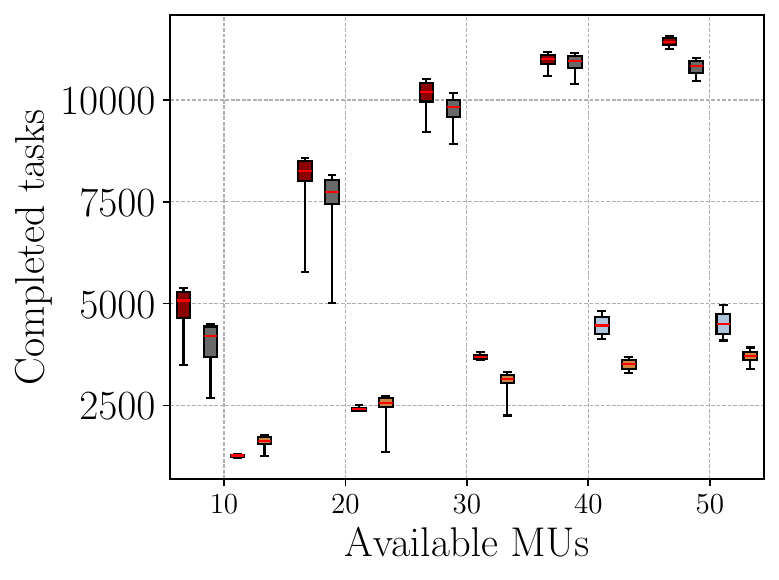}
       \label{fig:scenario_1_performance_comparison}
       }
        \subfloat[\footnotesize{Average collision ratio}]{
        \includegraphics[width=0.24\linewidth]{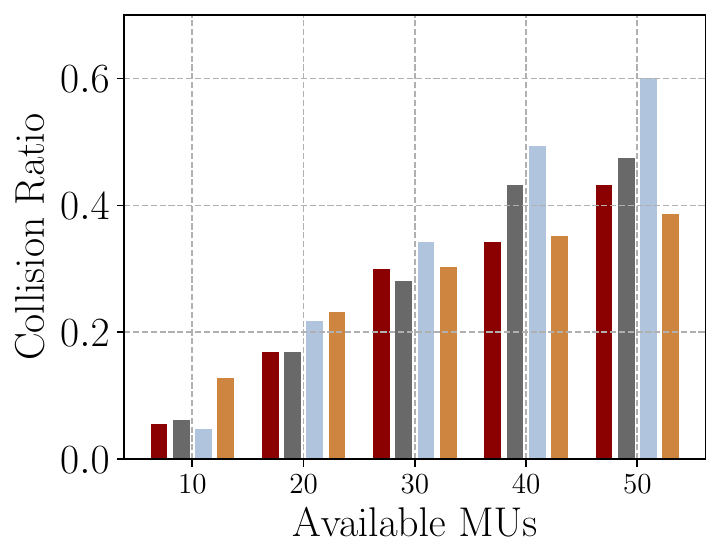}
        \label{fig:scenario_1_collisionRatio}
        }
        \subfloat[\footnotesize{Average completed tasks for each task type}]{
        \includegraphics[width=0.25\linewidth]{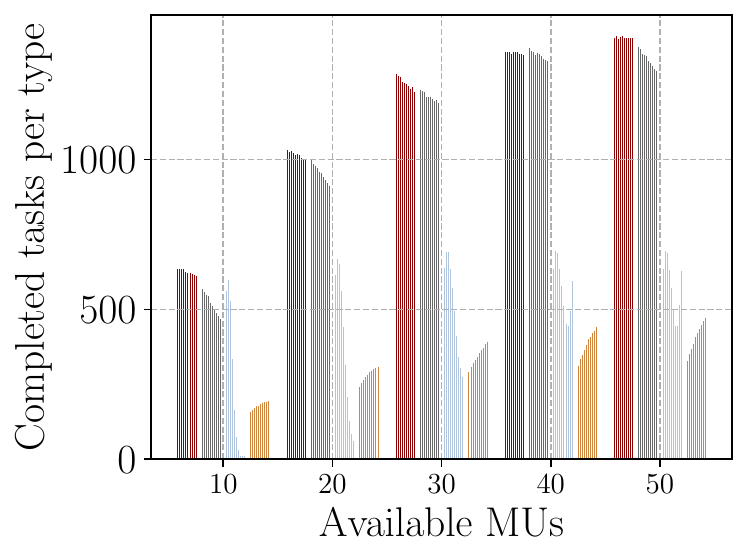}
        \label{fig:scenario_1_completed_tasks_per_task_type}
        }
        \subfloat[\footnotesize{Energy consumption}]{
        \includegraphics[width=0.23\linewidth]{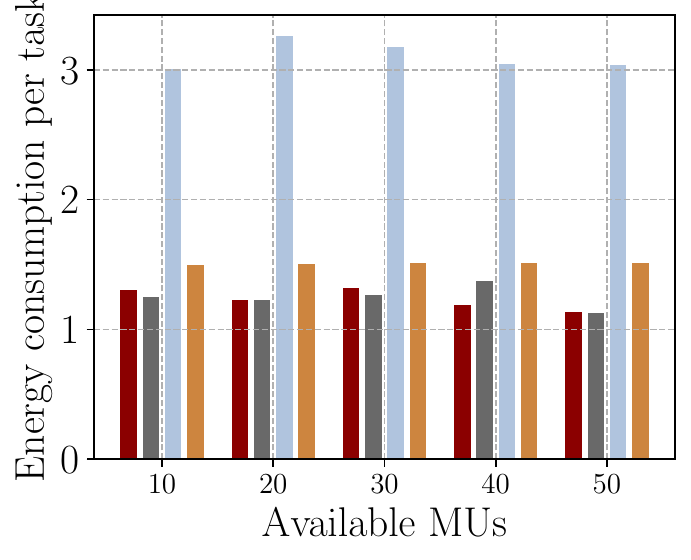}
        \label{fig:scenario_1_energy_consumption}
        }
    \caption{Scenario 2: Comparison of different performance metrics for varying number of available MUs}
    \vspace{-5mm}
\end{figure*}

In Fig.~\ref{fig:scenario_1_performance_comparison}, we compare the number of average weighted completed tasks within the considered time horizon.
The performance of every reference scheme improve as $K$ increases, as there are more available MUs which can potentially successfully perform the tasks.
Evidently, the performance of the proposed FDRL-PPO approach is consistently better than that of the reference schemes for all values of $K$.
FDRL-PPO outperforms the IPPO, the MOTP, and the RTPS by at least $4.2\%$, $60.6\%$, and $67.4\%$, respectively.
In Fig.~\ref{fig:scenario_1_collisionRatio}, we observe the effect of an increase in $K$ on the average collision ratio.
As the number of MUs increases, more MUs may be interested in performing the same tasks.
However, the task budget remains the same and thus, the average collisions increase as $K$ increases.
In RTPS, as the MUs are proposing to random tasks, the average collision ratio does not increase as rapidly as the reference schemes.
As a result, for $50$ MUs, the average collision ratio of RTPS is lower than our proposed approach, however, it manages to complete more weighted tasks on average than any other reference scheme.
The MUs learn faster by sharing their learning models and develop a task participation strategy which ensures task assignment.

In addition, the proposed FDRL-PPO is consistently fair in the execution of different task types for different number of MUs, as shown in Fig.~\ref{fig:scenario_1_completed_tasks_per_task_type}.
The behavior of RTPS is similar to the one observed in Fig. ~\ref{fig:basline_completed_tasks_per_tasktype}, where more difficult tasks where completed than easy ones.
On average, our FDRL-PPO performs all tasks fairly with respect to their task types in comparison to the rest of the reference algorithms which generally perform more easy tasks than the difficult ones.

In Fig. \ref{fig:scenario_1_energy_consumption}, we study the impact of a varying number of MUs on the normalized energy consumption w.r.t. the task completion ratio.
On average, FDRL-PPO outperforms IPPO, MOTP and RTPS by $9\%$, $133.3\%$ and $25\%$, respectively.
These gains are obtained thanks to the algorithm's ability to observe the impact of its decisions on the battery states of the MU.
Note that to maximize the income, an MUs must participate in as many tasks as possible, but to do that, it should have sufficient energy. 
Thus, the algorithm learns the impact of energy expenditure on the future states.

\textbf{\rc{Scenario 3 (Task load analysis):}} In this scenario, we study the effect of the number of available tasks per time step on the overall performance of the reference schemes.
For this, we vary the value of $N$ in the range $[10, 50]$ in steps of $10$ tasks, we set $K=15$ and we simulate for a time horizon of $1000$ time steps.

\begin{figure*}[t]
   \centering
       \begin{minipage}[c]{0.9\linewidth}
       \centering
       \includegraphics[width=0.6\linewidth]{figures/legend.pdf}
       \vspace*{-6mm}
       \end{minipage}
       \subfloat[\footnotesize{Average weighted completed tasks}]{
       \includegraphics[width=0.25\linewidth]{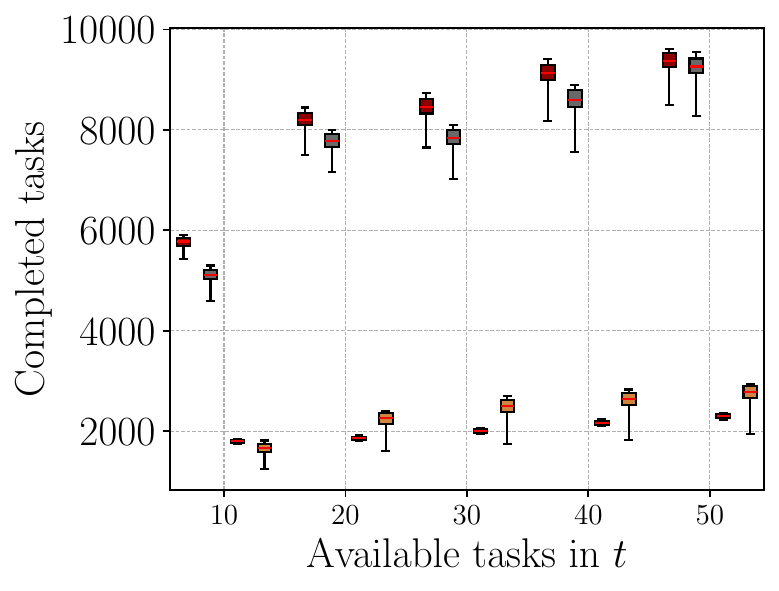}
       \label{fig:scenario_2_performance_comparison}
       }
        \subfloat[\footnotesize{Average collision ratio}]{
        \includegraphics[width=0.24\linewidth]{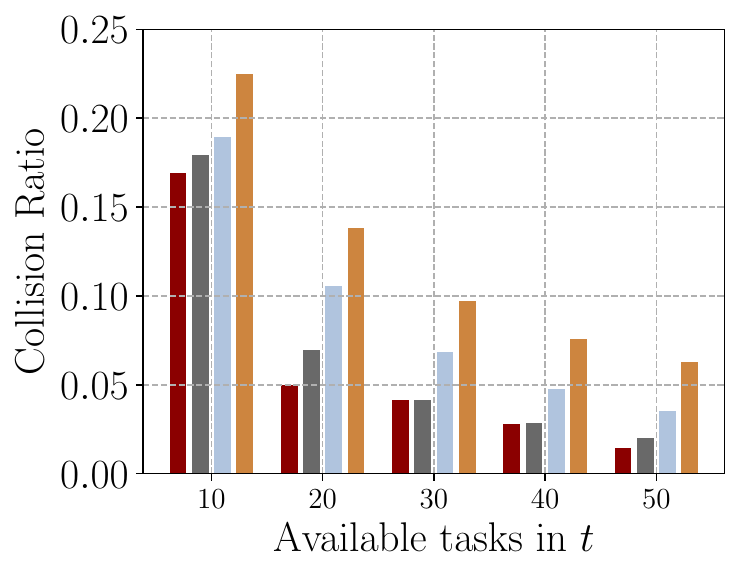}
        \label{fig:scenario_2_collisionRatio}
        }
        \subfloat[\footnotesize{Average completed tasks for each task type}]{
        \includegraphics[width=0.24\linewidth]{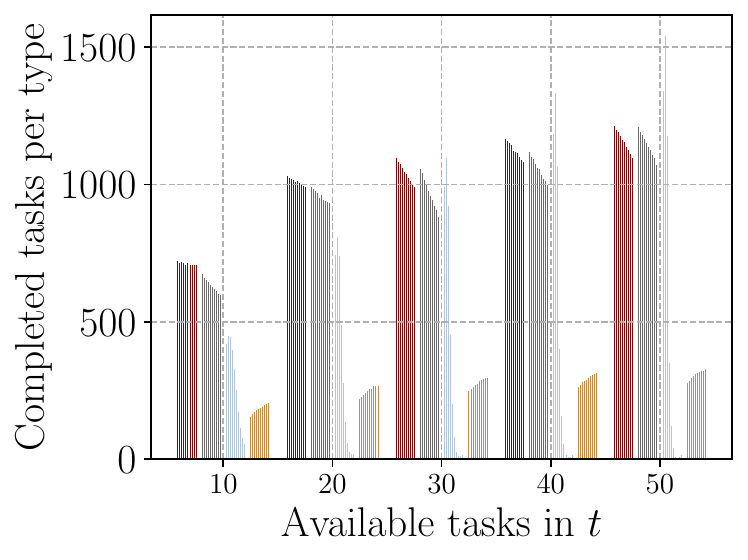}
        \label{fig:scenario_2_completed_tasks_per_task_type}
        }
        \subfloat[\footnotesize{Energy consumption}]{
        \includegraphics[width=0.23\linewidth]{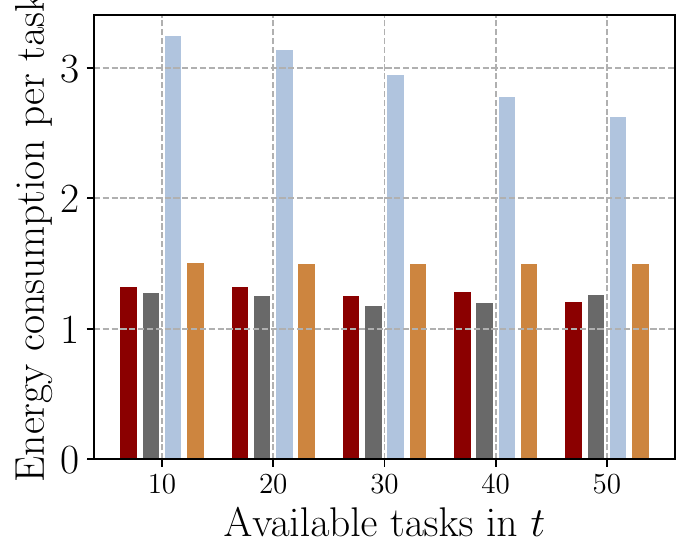}
        \label{fig:scenario_2_energy_consumption}
        }
        \caption{Scenario 3: Comparison of different performance metrics for varying number of available tasks per time step}
\vspace*{-7mm}
\end{figure*}
\begin{figure*}[t]
   \centering
       \subfloat[][\footnotesize{Average weighted completed tasks}]{
       \includegraphics[width=0.24\linewidth]{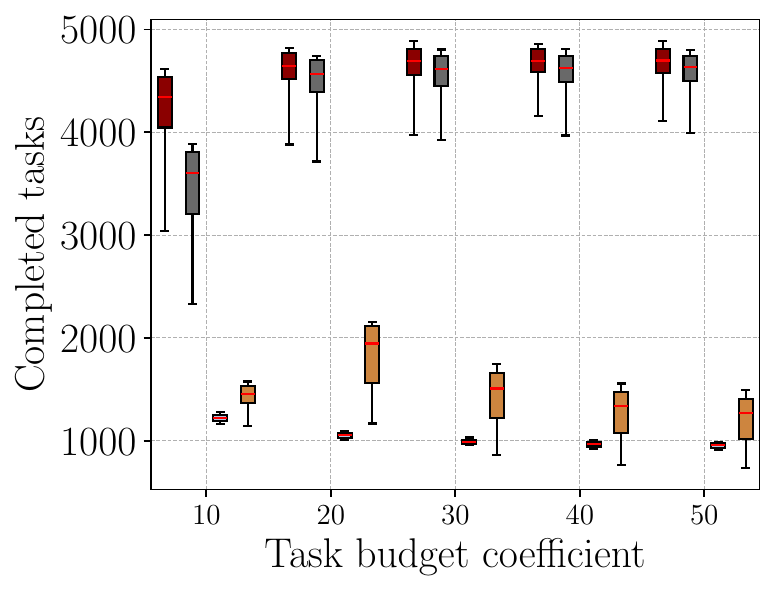}
       \label{fig:scenario_3_performance_comparison}
        }
        \subfloat[][\footnotesize{Average collision ratio}]{
        \includegraphics[width=0.24\linewidth]{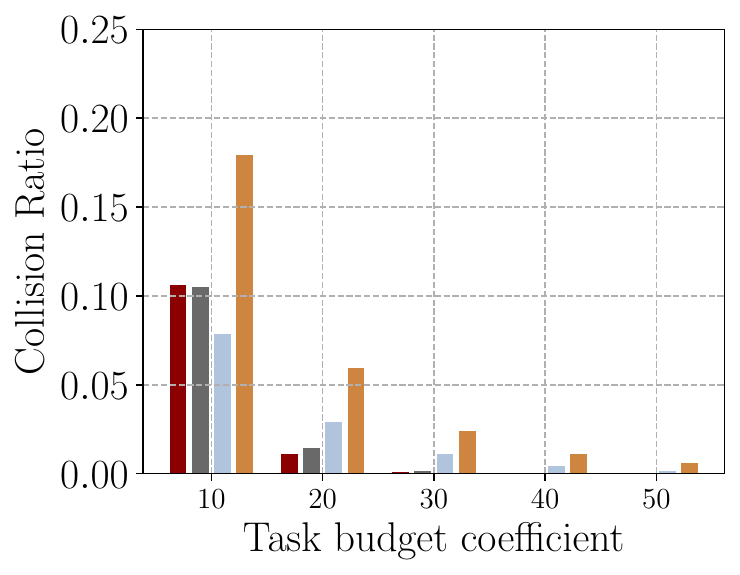}
        \label{fig:scenario_3_collisionRatio}
        }
       \subfloat[][\footnotesize{Average completed tasks for each task type}]{        
        \includegraphics[width=0.24\linewidth]{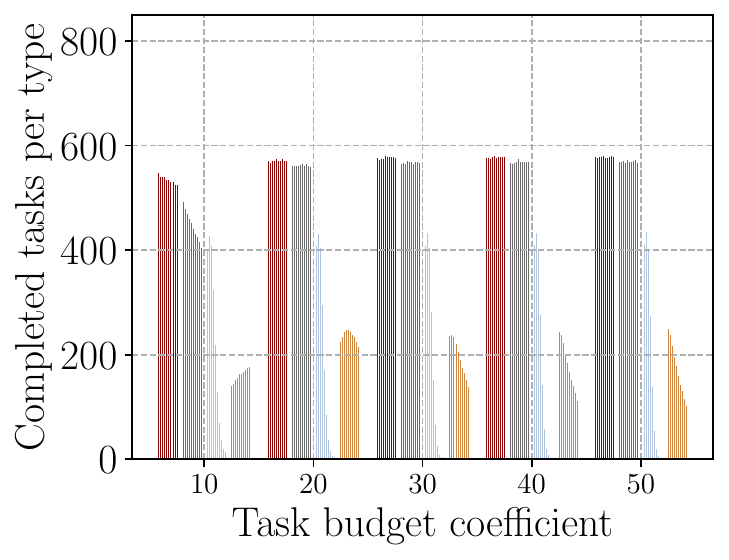}
        \label{fig:scenario_3_completed_tasks_per_task_type}
        }
        \subfloat[][\footnotesize{Energy consumption}]{        
        \includegraphics[width=0.23\linewidth]{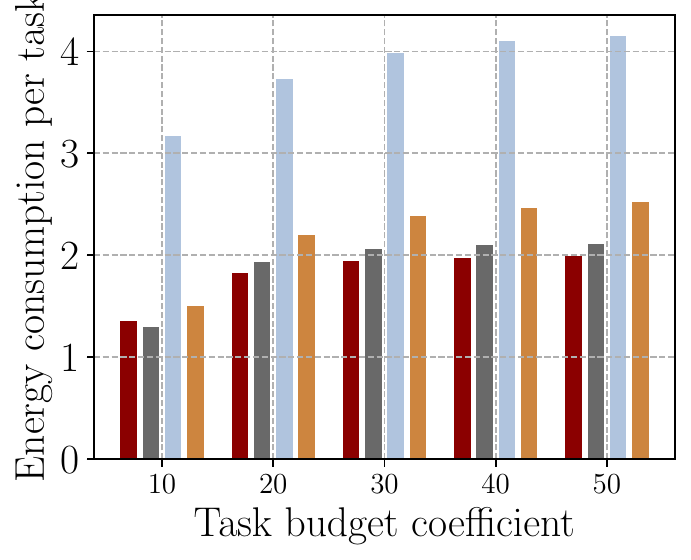}
        \label{fig:scenario_3_energy_consumption}
        }
    \caption{Scenario 4: Comparison of different performance metrics for varying task budget coefficient}
\end{figure*}

The performance comparison w.r.t number of average weighted completed tasks is showed in Fig.~\ref{fig:scenario_2_performance_comparison}.
As the number of available tasks in each time step increases, the MUs can choose from a larger set of tasks to which they can propose.
This also provides them better chances of getting accepted by the MCSP.
As a result, all the reference schemes perform better as the value of $N$ increases.
Our proposed FDRL-PPO algorithm always outperforms the IPPO algorithm.
Moreover, it is at least $70.6\%$ and $75.45\%$ better than the RTPS and the MOTP, respectively.
In FDRL-PPO, the MUs learn to participate in suitable tasks and leave other tasks such that potentially better suited MUs can participate and perform those tasks.
This is reflected also in the average collision ratio comparison in Fig.~\ref{fig:scenario_2_collisionRatio}.
Although the average collision ratio reduce as $N$ increases, our proposed FDRL-PPO achieves the smallest value as compared to all the reference algorithms.

In Fig.~\ref{fig:scenario_2_completed_tasks_per_task_type}, we analyze the fairness in task execution for increasing $N$.
We notice that when more tasks are available per time step, the proposed FDRL-PPO performs easier tasks more than the difficult ones.
The reference schemes also exhibit this behaviour.
However, FDRL-PPO still completes more difficult tasks than that of the references schemes as it learns to maximize the number of weighted completed tasks while considering the task difficulty.

In Fig.~\ref{fig:scenario_2_energy_consumption}, we investigate the impact of varying available tasks in a timesteps on the energy consumption per completed task for benchmark algorithms.
The proposed FDRL-PPO approach efficiently spends the available energy and outperforms the MOTP and RTPS algorithms by $116.7\%$ and $16.7\%$ respectively.
The IPPO algorithm and our proposed FDRL-PPO perform similar in terms of energy consumption as they both observe the impact of their actions on the battery state.

\textbf{\rc{Scenario 4 (Task budget analysis):}} In this scenario, we analyze the effect of changing the task budget coefficient.
Increasing the task budget implies that the MCSP can assign the task to more MUs.
For this scenario, we consider $K=10$ MUs and $N=10$ available tasks in each time step.
\begin{figure}[t]
   \centering
    \begin{minipage}[c]{0.49\linewidth}
        \includegraphics[width=\linewidth]{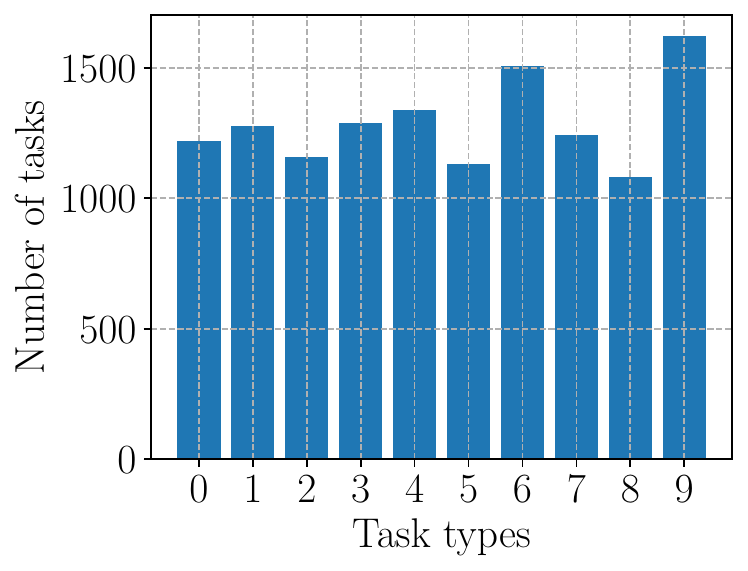}
        \vspace*{-7mm}
        \caption{Number of tasks per task type in the dataset}
        \label{fig:dataset_representation_jobtypeid}
    \end{minipage}
    \hfill
    \begin{minipage}[c]{0.48\linewidth}
        \includegraphics[width=\linewidth]{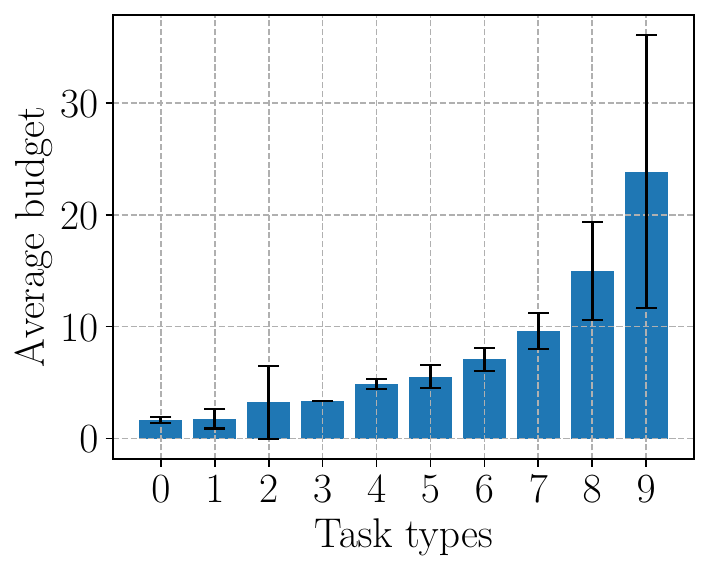}
        \vspace*{-7mm}
        \caption{Avg. budget per task type in the dataset}
        \label{fig:dataset_representation_reward}
    \end{minipage}
\end{figure}

In Fig.~\ref{fig:scenario_3_performance_comparison}, we observe that the number of average weighted completed tasks of the proposed FDRL-PPO approach saturates as the task budget coefficient $\eta$ increases.
This is because the task participation strategy of the FDRL-PPO algorithm takes task budget into account.
In this sense, having higher task budget only reduces the collisions, which can be observed in Fig.~\ref{fig:scenario_3_collisionRatio}.
Similar trend is observed in case of IPPO scheme, however, the proposed FDRL-PPO approach is at least $1.5\%$ and at most $20.4\%$ better than the IPPO.
The performance of MOTP reduces and stabilizes to a value which is $79.71\%$ worse than our proposed FDRL-PPO.
This is because MOTP prefers difficult tasks more than the easy tasks and this already involves collisions and performance reduction.
However, as the task budget coefficient increases, the task budget increases which allows more MUs to be accepted for the same task.
Due to this, the collisions in MOTP reduce with increasing $\eta$, however, the average weighted completed tasks do not increase.
This is because more MUs perform the same tasks.
The performance of the RTPS improves first since potentially more MUs can be assigned to task execution.
However, as the task budget coefficient increases further, more MUs are 
assigned for task execution, however, most of them are unable to successfully complete the task.
This results in a lot of wasted energy without any monetary benefit for the MUs since the tasks on average are not completed.
Because higher task budget only benefits in more task assignments, however, the uncertainty in the environment and random resource allocation by the RTPS affects its performance.
This is visible in Fig.~\ref{fig:scenario_3_completed_tasks_per_task_type}, where the RTPS completes more difficult tasks for $\eta=10$, however, as $\eta$ increases, RTPS displays a trend of task completion similar to other reference schemes.
For the proposed FDRL-PPO algorithm along with the IPPO, the task execution becomes fair as the budget increases.
In Fig.~\ref{fig:scenario_3_energy_consumption},
we see the average energy spent by the MU in performing a task successfully.
As the task budget increases, more MUs participate in the task execution and therefore, the overall average energy spent per executed task increases.
Our proposed FDRL-PPO outperforms all the benchmark approaches consistently in this scenario.
The FDRL-PPO and the IPPO again spend considerably similar energy per task.
However, as the task budget increases, more MUs perform the task and receive more information about the different states through federation.
As a result, the FDRL-PPO eventually outperforms IPPO for higher task budget coefficients by at least $5\%$.
FDRL-PPO outperforms MOTP by at least $100\%$.
As the task budget coefficient increases, the task budget increases, allowing RTPS to perform more tasks as the MUs fit within the budget.
In this case, the energy expenditure of the RTPS is at least $10\%$ worse than our proposed FDRL-PPO algorithm.
\vspace*{-4mm}
\subsection{Evaluation using a Real-World Dataset}
\label{sec:dataset_evaluation}
We showcase the performance of our approach with a real-world dataset, acquired through a well-known MCS-application which users can download on Apple AppStore or Google PlayStore.\footnote{The datasets used in our publication are not publicly available, as the company behind the MCS-application did not authorize public usage. Still, interested readers may get access to the datasets by contacting the authors.}
Upon launching the app, users registered with their personal data. 
Similarly, data requesters registered with the app to create sensing tasks with their description, requirements, and their geographical positions. 
Hence, a two-sided market evolved on the platform, matching users and jobs. 
In this case, the jobs correspond to the tasks and the users to the MUs. 
We view the application platform as the MCSP.
Every task has geographical coordinates, a task type identifier, and a reward. 
We select a geographical target area covering a major metropolitan German city.

In Fig.~\ref{fig:dataset_representation_jobtypeid} and Fig.~\ref{fig:dataset_representation_reward}, we analyze the distribution of the available tasks in the dataset according to their task types and also their average monetary budget.
The dataset features $526$ different task descriptions such as taking pictures of some specific places, confirming the existence of certain speed cameras on the highways, or simple sensing tasks such as noise level sensing.
Each of the task had a different task budget.
To reduce the granularity in the dataset, we sorted the tasks according to their budget and divided them into $10$ different task types.
In Fig.~\ref{fig:dataset_representation_jobtypeid}, we see the number of tasks from each task type.
Although not strictly uniform, the number of tasks in each task type are similar, however, their arrival process is non-uniform.
This means, for example, there could be only easy tasks available for several time steps in the dataset.
This differs from the synthetically generated dataset which contains uniformly distributed tasks with respect to their task types.
In Fig.~\ref{fig:dataset_representation_reward}, we see the average task budget for every task of each task type.
This distribution is exponentially increasing in contrast to linear increase in the synthetically generated data.

\begin{figure*}[t]
   \centering
    \begin{minipage}[c]{\linewidth}
    \centering
    \includegraphics[width=0.6\linewidth]{figures/legend.pdf}
    \vspace*{-6mm}
    \end{minipage}
    \subfloat[\scriptsize{Average weighted completed tasks}]{
       \includegraphics[width=0.25\linewidth]{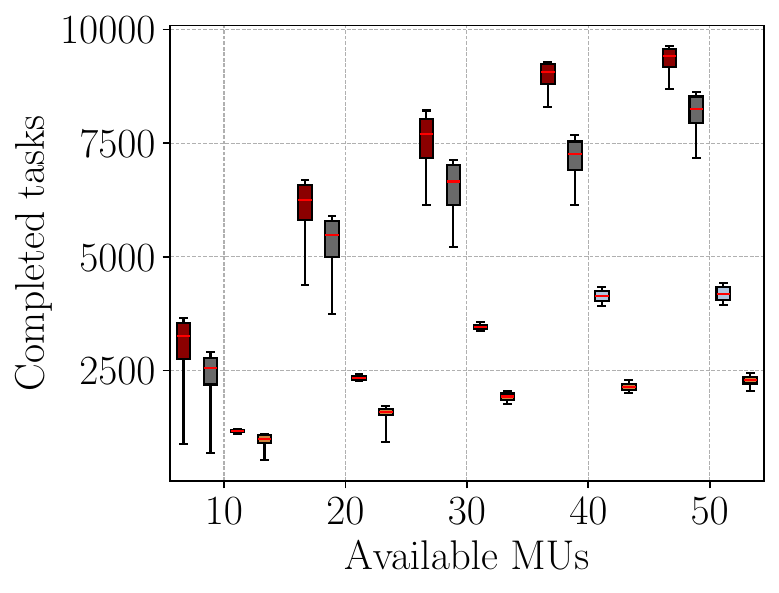}
       \label{fig:scenario_1_performance_comparison_dataset}
       }
    \subfloat[\scriptsize{Average collision ratio}]{
        \includegraphics[width=0.24\linewidth]{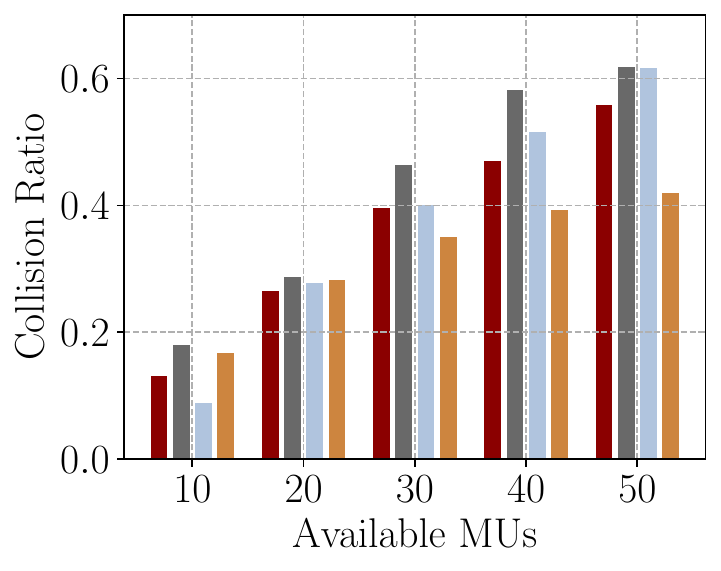}
        \label{fig:scenario_1_collisionRatio_dataset}
        }
    \subfloat[\scriptsize{Average completed task per task type}]{
        \includegraphics[width=0.25\linewidth]{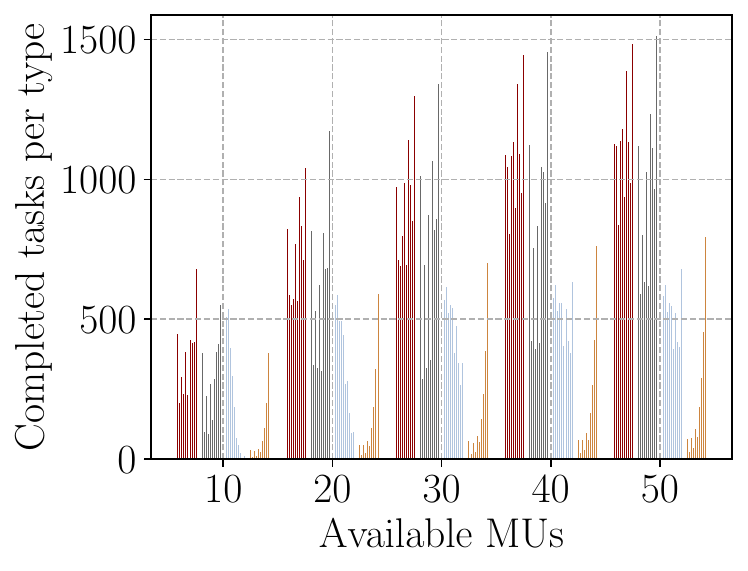}
        \label{fig:scenario_1_completedTasksPerTaskType_dataset}
        }
    \subfloat[\scriptsize{Energy consumption}]{
        \includegraphics[width=0.23\linewidth]{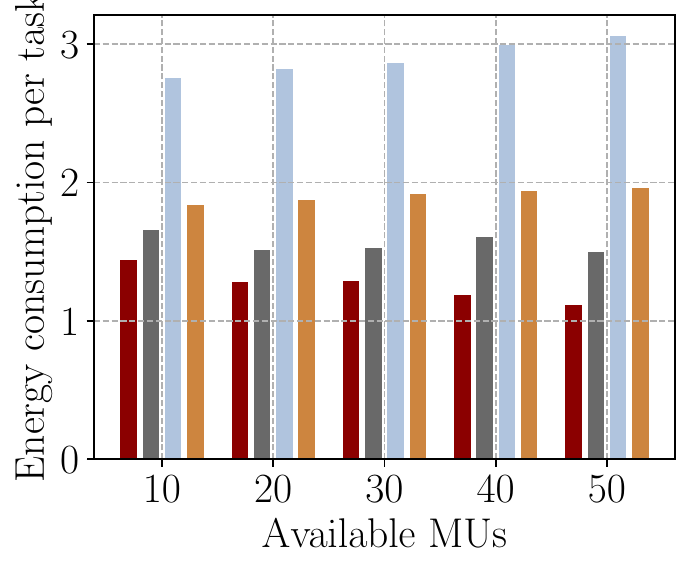}
        \label{fig:scenario_1_energyConsumption_dataset}
        }
    \caption{Scenario 2: Comparison of different performance metrics for varying number of available MUs}
    \label{fig:dataset_scenario_1}
    \vspace{-5mm}
\end{figure*}

\rc{In Fig.~\ref{fig:dataset_scenario_1}, we consider Scenario 2 (MU scalability analysis) from Section \ref{sec:simulation_results}.}
We change the number of MUs in the range $[10, 50]$ and keep the number of available tasks per time step, $N=15$.
Based on the dataset, we simulated for $T=850$ time steps.
While comparing the performance in Fig.~\ref{fig:scenario_1_performance_comparison_dataset}, we see that as the number of MUs increases, the performance increases.
This result is similar to the simulation results obtained in Scenario 2 in Section \ref{sec:simulation_results}.
Our FDRL-PPO performs at least $21.78\%$, $64.06\%$, and $69.56\%$ better than IPPO, MOTP, and RTPS, respectively.
In Fig. \ref{fig:scenario_1_collisionRatio_dataset}, the average number of collisions increase as the number of available MUs increase.
However, our proposed FDRL-PPO approach on average has fewer collisions than IPPO and MOTP algorithms.
The collisions in RTPS are the lowest due to the randomness.
In Fig.~\ref{fig:scenario_1_completedTasksPerTaskType_dataset}, we compare average completed tasks per task type.
As the number of MUs increase, more MUs are available to perform the available tasks.
Our proposed FDRL-PPO consistently performs more tasks in each type.
The shape of the bars eventually look like the task type distribution of the dataset, which confirms that our approach fairly performs the tasks of each type.
In Fig.~\ref{fig:scenario_1_energyConsumption_dataset}, our approach consumes less energy per completed task than all of the benchmark approaches.
It outperforms IPPO, MOTP, and RTPS by at least $14.2\%$, $100\%$, and $28.6\%$ respectively.

\begin{figure*}[t]
   \centering
    \subfloat[\scriptsize{Average weighted completed tasks}]{
       \includegraphics[width=0.25\linewidth]{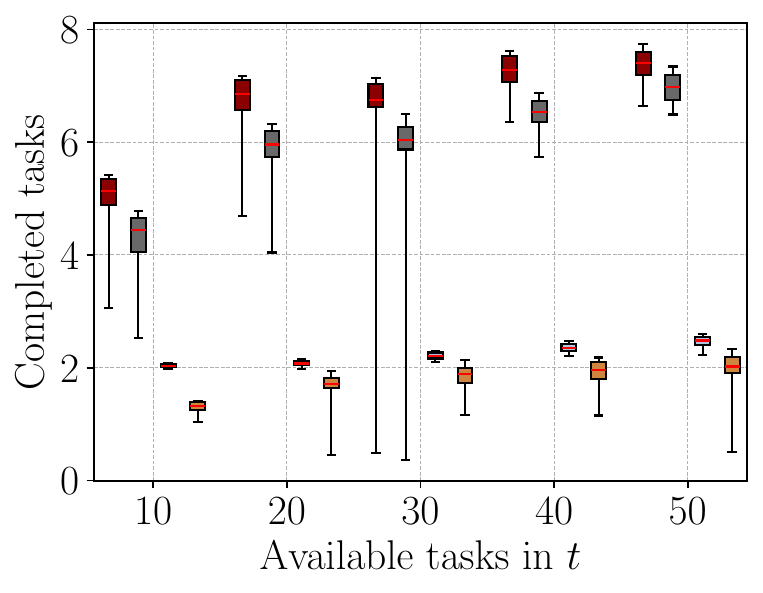}
       \label{fig:dataset_scenario_2_performance_comparison}
       }
    \subfloat[\scriptsize{Average collision ratio}]{
        \includegraphics[width=0.24\linewidth]{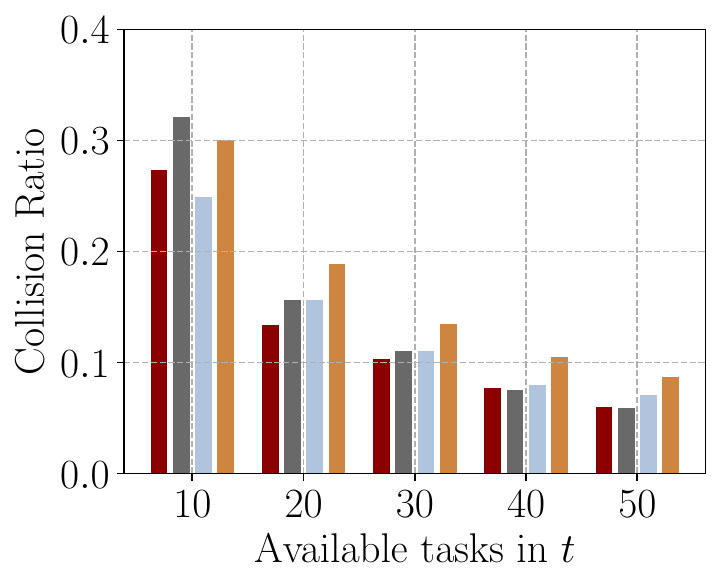}
        \label{fig:dataset_scenario_2_collisionRatio}
        }
    \subfloat[\scriptsize{Average completed tasks per task type}]{
        \includegraphics[width=0.25\linewidth]{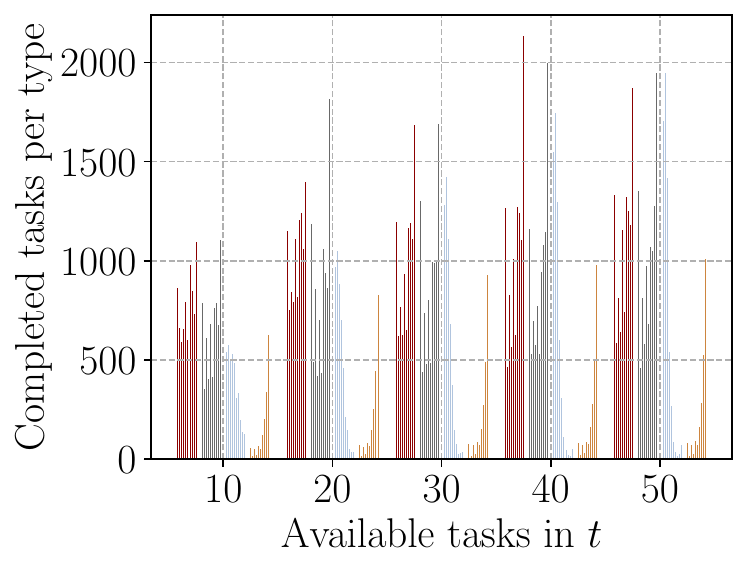}
        \label{fig:dataset_scenario_2_completedTasksPerTaskType}
        }
    \subfloat[\scriptsize{Energy consumption}]{
        \includegraphics[width=0.23\linewidth]{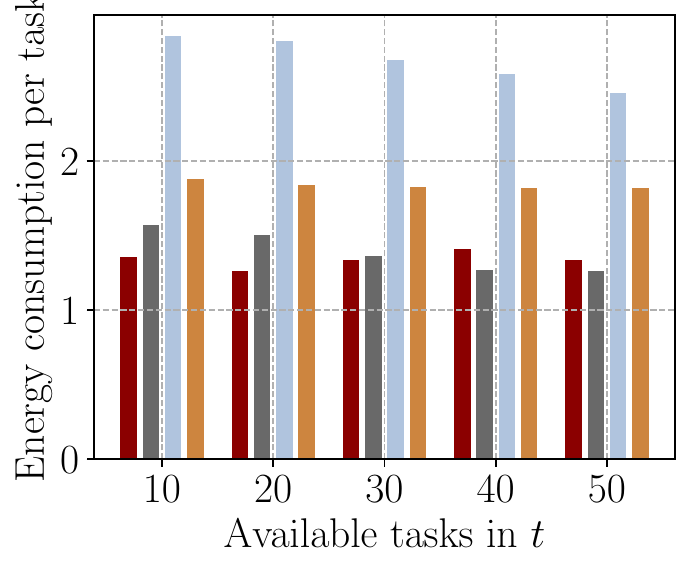}
        \label{fig:dataset_scenario_2_energyConsumption}
        }
    \caption{Scenario 3: Comparison of different performance metrics for varying number of available tasks per time step}
    \label{fig:dataset_scenario_2}
    \vspace{-7mm}
\end{figure*}

\begin{figure*}[t]
    \centering
    \subfloat[\scriptsize{Average weighted completed tasks}]{
       \includegraphics[width=0.25\linewidth]{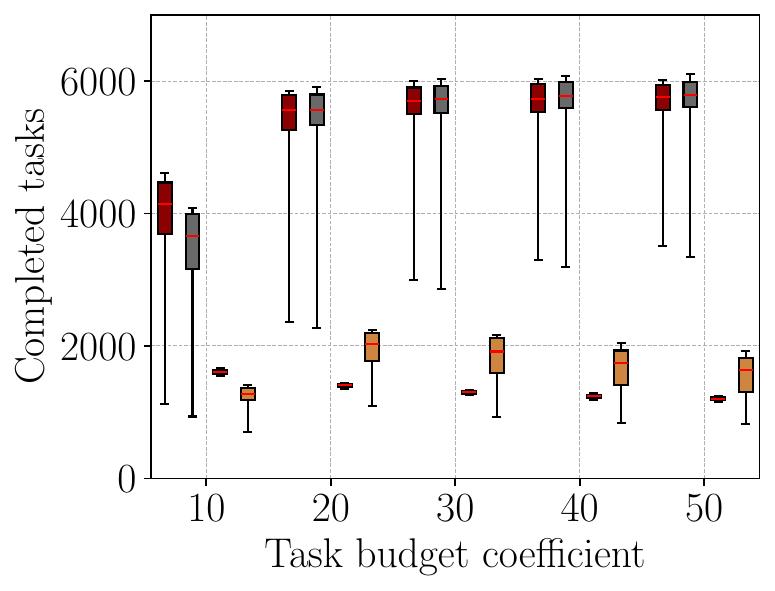}
       \label{fig:dataset_scenario_3_performance_comparison}
    }
    \subfloat[\scriptsize{Average collision ratio}]{
        \includegraphics[width=0.24\linewidth]{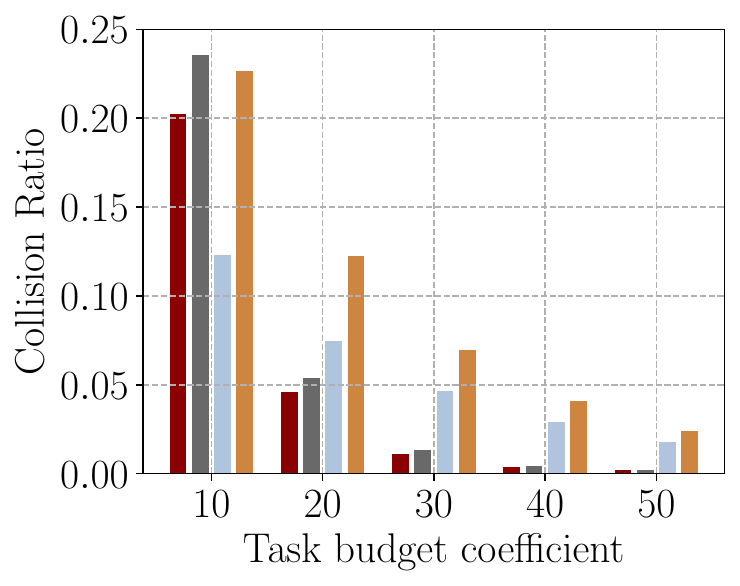}
        \label{fig:dataset_scenario_3_collisionRatio}
    }
    \subfloat[\scriptsize{Average completed task per task type}]{
        \includegraphics[width=0.25\linewidth]{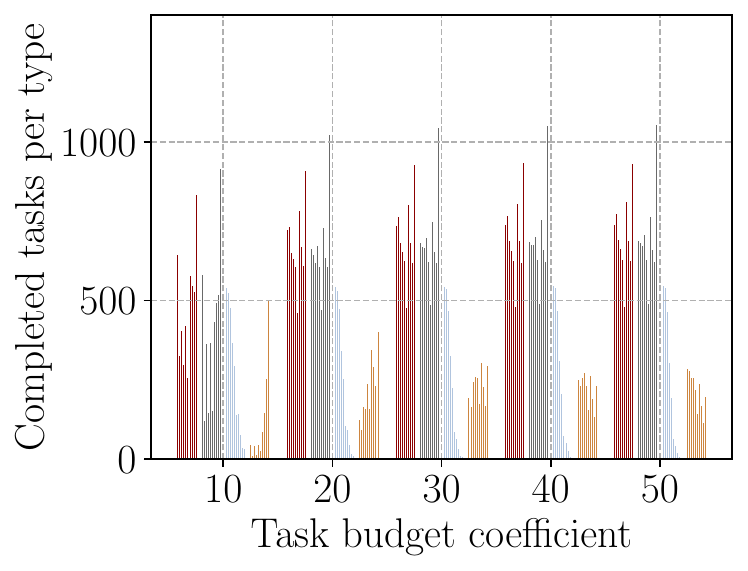}
        \label{fig:dataset_scenario_3_completedTasksPerTaskType}
    }
    \subfloat[\scriptsize{Energy consumption}]{
        \includegraphics[width=0.24\linewidth]{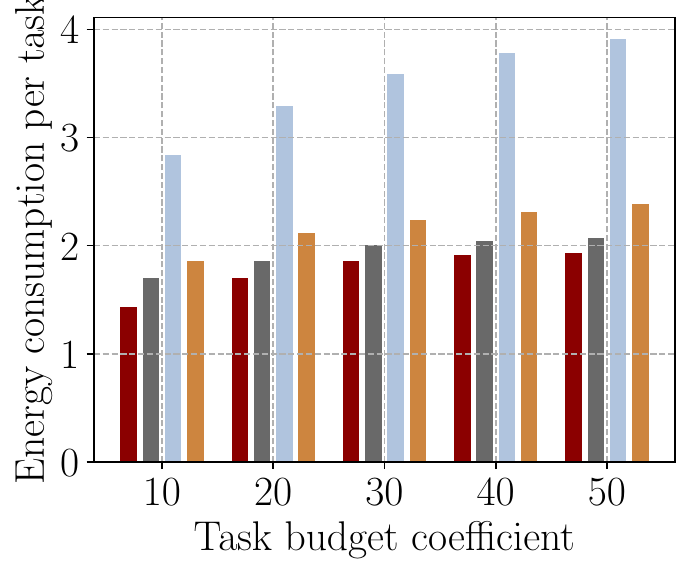}
        \label{fig:dataset_scenario_3_energyConsumption}
    }
    \caption{Scenario 4: Comparison of different performance metrics for varying task budget coefficient}
    \label{fig:dataset_scenario_3}
\end{figure*}

\rc{In Fig.~\ref{fig:dataset_scenario_2}, we employ Scenario 3 (Task load analysis) from Section \ref{sec:simulation_results} where we consider $K=15$ MUs.}
We vary the number $N$ of available tasks in every time step in the range $[10, 50]$.
Due to finite number of tasks in the dataset, the length of the time horizon was changed in the range from $T=1280$ timesteps to  $T=256$ timesteps as the number $N$ of available tasks increase.
This is why the performance in Fig.~\ref{fig:dataset_scenario_2_performance_comparison} is measured in average weighted completed tasks in each time step.
As the number of available tasks in each time step increases, the MUs have more options to choose a better task.
Thus, we see that the performance improves.
This is evident in Fig.~\ref{fig:dataset_scenario_2_collisionRatio} where the collisions decrease with increasing number of available tasks.
In this scenario, our proposed FDRL-PPO algorithm outperforms IPPO, MOTP, and RTPS by atleast $5.64\%$, $66.47\%$, and $72.70\%$, respectively.
In Fig.~\ref{fig:dataset_scenario_2_completedTasksPerTaskType}, we compare average completed tasks per task type.
As the number of available tasks increase, the MUs can choose from a larger set of tasks which reduces the collisions.
Our proposed FDRL-PPO consistently performs more tasks in each type than the benchmark algorithms.
In Fig.~\ref{fig:dataset_scenario_2_energyConsumption}, our approach consumes less energy per completed task than all of the benchmark approaches.
The energy consumption of IPPO and FDRL-PPO is similar as the algorithms cautiously spend the available energy by considering the impact on the future timesteps.
Our proposed FDRL-PPO outperforms IPPO, MOTP, and RTPS by $9.2\%$, $92.3\%$, and $38.5\%$ on average, respectively.

\rc{In Fig.~\ref{fig:dataset_scenario_3}, we analyze the effect of the task budget coefficient $\eta$ on the tasks from the dataset based on Scenario 4 (Task budget analysis).}
We used $K=10$ MUs and $N=10$ available tasks in every time step.
In Fig.~\ref{fig:dataset_scenario_3_performance_comparison}, we see that in a constrained task budget coefficient value $\eta=10$, our algorithm outperforms the IPPO, MOTP, and RTPS by $11.58\%$, $61.14\%$, and $69.16\%$, respectively.
However, for higher $\eta$ values, the performance of the IPPO scheme improves significantly.
There are multiple reasons for this behaviour.
First, the task arrival in the dataset is non-uniform.
Thus, the results in the simulations for Scenario 4 are different than the dataset evaluation of Scenario 4.
Second, for high $\eta$ values, the MCSP has higher budget for the same tasks.
Consequently, more MUs will be accepted at the MCSP for performing the tasks.
As a result, the FDRL-PPO performs similar to the IPPO scheme.
We also see this in Fig.~\ref{fig:dataset_scenario_3_collisionRatio}, where the average collision ratio drops drastically in case of FDRL-PPO and IPPO.
With increased budget for every task, the FDRL-PPO and IPPO converge however, the MOTP and RTPS struggle to perform well due to collisions and failed task completions.
In Fig.~\ref{fig:dataset_scenario_3_completedTasksPerTaskType}, we see that the average number of tasks performed per task type increase as the task budget coefficient increases.
Our proposed FDRL-PPO algorithm performs more tasks in each type on average than the benchmark algorithms.
In terms of the energy consumption per completed task in Fig. \ref{fig:dataset_scenario_3_energyConsumption}, our proposed approach consistently outperforms the IPPO, MOTP, and RTPS by at least $10.5\%$, $105.2\%$, and $21\%$ respectively.
This result confirms the simulations and showcases the advantages of a federated approach in the distributed learning setup.

\vspace*{-3mm}
\section{Conclusion}
\label{sec:conclusion}
In this work, we have studied the task participation problem of the energy harvesting MUs in a budgeted and location aware MCS system.
By considering their own preferences and their available resources, the MUs make task participation decisions and convey them to the MCSP.
The MCSP then chooses the MUs according to the task budget.
The optimization-based approaches require perfect non-causal information about the MCS system to maximize the average weighted completed tasks.
However, availability of such information at the MUs or at the MCSP is unrealistic in practical MCS systems.
Thus, to find an efficient and reliable task participation strategy for the MUs to maximize their income under incomplete information is a challenge.
To this aim, we have proposed a FDRL-PPO approach which solves the task participation problem of the MUs without the unrealistic requirement of perfect non-causal information.
Using FDRL-PPO, the MUs learn to propose to suitable tasks over time based on their preferences and available resources.
The MUs are battery operated and replenish their batteries using energy harvesting.
As a result, the amount of available energy in MUs' battery affects their availability to perform tasks and obstructs their learning.
To remedy this, the proposed approach utilizes the federated architecture in which the MUs share their learned models with other MUs such that they can collectively find an efficient and reliable task participation strategy.
This makes the proposed approach robust to MU drop-outs and join-ins.
We have performed extensive numerical evaluations to compare the performance of our proposed approach with the reference algorithms using synthetic as well as real-world datasets.
The simulation results validated the superior performance of the FDRL-PPO approach in various considered scenarios.

\begin{spacing}{0.87}
\bibliographystyle{./bibliography/IEEEtran}
\bibliography{./bibliography/IEEEabrv,./bibliography/sample}

@IEEEtranBSTCTL{IEEEexample:BSTcontrol,
  CTLuse_forced_etal       = "yes",
  CTLmax_names_forced_etal = "4",
  CTLnames_show_etal       = "4" 
}

@article{sterz2022multi,
  title={Multi-Stakeholder Service Placement via Iterative Bargaining With Incomplete Information},
  author={Sterz, Artur and Felka, Patrick and Simon, Bernd and Klos, Sabrina and Klein, Anja and Hinz, Oliver and Freisleben, Bernd},
  journal={IEEE/ACM Trans. on Netw.},
  volume={30},
  number={4},
  pages={1822--1837},
  year={2022},
  publisher={IEEE}
}

@article{akerlof1970market,
  title={The Market for Lemons: Quality Uncertainty and the Market Mechanism},
  author={Akerlof, George A},
  journal={Quarterly Journal of Economics},
  volume={84},
  pages={488--500},
  year={1970}
}

@article{bergh2019information,
  title={Information asymmetry in management research: Past accomplishments and future opportunities},
  author={Bergh, Donald D and Ketchen Jr, David J and Orlandi, Ilaria and Heugens, Pursey PMAR and Boyd, Brian K},
  journal={Journal of management},
  volume={45},
  number={1},
  pages={122--158},
  year={2019},
  publisher={SAGE Publications Sage CA: Los Angeles, CA}
}

@article{jiang2016share,
  title={To share or not to share: Demand forecast sharing in a distribution channel},
  author={Jiang, Baojun and Tian, Lin and Xu, Yifan and Zhang, Fuqiang},
  journal={Marketing Science},
  volume={35},
  number={5},
  pages={800--809},
  year={2016},
  publisher={INFORMS}
}

@misc{schulman2017proximal,
      title={Proximal Policy Optimization Algorithms}, 
      author={John Schulman and Filip Wolski and Prafulla Dhariwal and Alec Radford and Oleg Klimov},
      year={2017},
      eprint={1707.06347},
      archivePrefix={arXiv},
      primaryClass={cs.LG}
}

@online{Foursquare,
  title = {FourSquare},
  url = {https://location.foursquare.com/},
  urldate = {2023-05-10}
}

@online{komoot,
  title = {Komoot},
  url = {https://www.komoot.com/},
  urldate = {2023-05-10}
}

@ARTICLE{OPAT_Huang_2022,
  author={Huang, Yang and Chen, Honglong and Ma, Guoqi and Lin, Kai and Ni, Zhichen and Yan, Na and Wang, Zhibo},
  journal={IEEE Trans. on Industrial Informatics}, 
  title={{OPAT}: Optimized Allocation of Time-Dependent Tasks for Mobile Crowdsensing},
  year={2022},
  volume={18},
  number={4},
  pages={2476-2485},
  doi={10.1109/TII.2021.3094527}}

@ARTICLE{Xu_Task_allocation_DRL_2023,
  author={Xu, Chenghao and Song, Wei},
  journal={IEEE Trans. on Netw. Sci. and Engg.}, 
  title={Intelligent Task Allocation for Mobile Crowdsensing With Graph Attention Network and Deep Reinforcement Learning}, 
  year={2023},
  volume={10},
  number={2},
  pages={1032-1048},
  doi={10.1109/TNSE.2022.3226422}}

@ARTICLE{Spectrum_sensing_MCS_2021,
  author={Dong, Xuewen and You, Zhichao and Luan, Tom H. and Yao, Qingsong and Shen, Yulong and Ma, Jianfeng},
  journal={IEEE IoT Journal}, 
  title={Optimal Mobile Crowdsensing Incentive Under Sensing Inaccuracy}, 
  year={2021},
  volume={8},
  number={10},
  pages={8032-8043},
  doi={10.1109/JIOT.2020.3042979}}

@INPROCEEDINGS{Dinh_environmental_monitoring_2022,
  author={Dinh, Tuan Anh Nguyen and Nguyen, Anh Duy and Nguyen, Truong Thao and Nguyen, Thanh Hung and Nguyen, Phi Le},
  booktitle={IEEE Wireless Commun. and Networking Conf. (WCNC)}, 
  title={Spatial-temporal Coverage Maximization in Vehicle-based Mobile Crowdsensing for Air Quality Monitoring}, 
  year={2022},
  volume={},
  number={},
  pages={1449-1454},
  doi={10.1109/WCNC51071.2022.9771711}}

@INPROCEEDINGS{Bernd_ICC_2022,
  author={Simon, Bernd and Dongare, Sumedh and Mahn, Tobias and Ortiz, Andrea and Klein, Anja},
  booktitle={IEEE Int. Conf. on Commun.}, 
  title={Delay- and Incentive-Aware Crowdsensing: A Stable Matching Approach for Coverage Maximization}, 
  year={2022},
  volume={},
  number={},
  pages={2984-2989},
  doi={10.1109/ICC45855.2022.9838603}}

@INPROCEEDINGS{Dongare_EHMCS_2022,
  author={Dongare, Sumedh and Ortiz, Andrea and Klein, Anja},
  booktitle={IEEE Global Commun. Conf.}, 
  title={Deep Reinforcement Learning for Task Allocation in Energy Harvesting Mobile Crowdsensing}, 
  year={2022},
  volume={},
  number={},
  pages={269-274},
  doi={10.1109/GLOBECOM48099.2022.10001204}}

@ARTICLE{MCS_intro_Dai_21,
  author={Dai, Chenxin and Wang, Xiumin and Liu, Kai and Qi, Deyu and Lin, Weiwei and Zhou, Pan},
  journal={IEEE Trans. on Mobile Comput.}, 
  title={Stable Task Assignment for Mobile Crowdsensing With Budget Constraint}, 
  year={2021},
  volume={20},
  number={12},
  pages={3439-3452},
  doi={10.1109/TMC.2020.3000234}}

@ARTICLE{Ganti2011MCS_intro,
  author={Ganti, Raghu K. and Ye, Fan and Lei, Hui},
  journal={IEEE Commun. Mag.}, 
  title={Mobile crowdsensing: current state and future challenges}, 
  year={2011},
  volume={49},
  number={11},
  pages={32-39},
  doi={10.1109/MCOM.2011.6069707}}

@INPROCEEDINGS{Pryss2018mHealth,
  author={Pryss, Rüdiger and Schobel, Johannes and Reichert, Manfred},
  booktitle={Int. Workshop on Requirements Engineering for Self-Adaptive, Collaborative, and Cyber Physical Systems (RESACS)}, 
  title={Requirements for a Flexible and Generic {API} Enabling Mobile Crowdsensing mHealth Applications}, 
  year={2018},
  volume={},
  number={},
  pages={24-31},
  doi={10.1109/RESACS.2018.00010}}

@INPROCEEDINGS{iot_mcs_Jian_2015,
  author={An, Jian and Gui, Xiaolin and Yang, Jianwei and Yu, Sun and He, Xin},
  booktitle={IEEE Int. Conf. on Multimedia Big Data}, 
  title={Mobile Crowd Sensing for Internet of Things: A Credible Crowdsourcing Model in Mobile-Sense Service}, 
  year={2015},
  volume={},
  number={},
  pages={92-99},
  doi={10.1109/BigMM.2015.62}}

@ARTICLE{EH-WSNs_sandhu_2021,
  author={Sandhu, Muhammad Moid and Khalifa, Sara and Jurdak, Raja and Portmann, Marius},
  journal={IEEE IoT Journal}, 
  title={Task Scheduling for Energy-Harvesting-Based IoT: A Survey and Critical Analysis}, 
  year={2021},
  volume={8},
  number={18},
  pages={13825-13848},
  doi={10.1109/JIOT.2021.3086186}}

@INPROCEEDINGS{Dongare_Globecom_2023,
  author={Dongare, Sumedh and Ortiz, Andrea and Klein, Anja},
  booktitle={IEEE Global Commun. Conf.}, 
  title={Federated Deep Reinforcement Learning for Task Participation in Mobile Crowdsensing}, 
  year={2023},
  volume={},
  number={},
  pages={4436-4441},
  doi={10.1109/GLOBECOM54140.2023.10436786}}

@INPROCEEDINGS{C1:own:Dongare2024b,
  author={Dongare, Sumedh and Simon, Bernd and Ortiz, Andrea and Klein, Anja},
  booktitle={IEEE Int. Conf. on Commun.}, 
  title={Two-Sided Learning: A Techno-Economic View of Mobile Crowdsensing under Incomplete Information}, 
  year={2024},}

@ARTICLE{Unified_JSCC_XLi_2023,
  author={Li, Xiaoqian and Feng, Gang and Sun, Yao and Qin, Shuang and Liu, Yijing},
  journal={IEEE Trans. on Mob. Comput.}, 
  title={A Unified Framework for Joint Sensing and Communication in Resource Constrained Mobile Edge Networks}, 
  year={2023},
  volume={22},
  number={10},
  pages={5643-5656},
  doi={10.1109/TMC.2022.3188804}
}

@ARTICLE{Decentralized_TA_MARL_Xu_2023,
  author={Xu, Chenghao and Song, Wei},
  journal={IEEE IoT Journal}, 
  title={Decentralized Task Assignment for Mobile Crowdsensing With Multi-Agent Deep Reinforcement Learning}, 
  year={2023},
  volume={10},
  number={18},
  pages={16564-16578},
  doi={10.1109/JIOT.2023.3268846}
}

@ARTICLE{Personalized_task_oriented_Wang_2021,
  author={Wang, Zhibo and Zhao, Jing and Hu, Jiahui and Zhu, Tianqing and Wang, Qian and Ren, Ju and Li, Chao},
  journal={IEEE Trans. on Mob. Comput.}, 
  title={Towards Personalized Task-Oriented Worker Recruitment in Mobile Crowdsensing}, 
  year={2021},
  volume={20},
  number={5},
  pages={2080-2093},
  doi={10.1109/TMC.2020.2973990}}

@INPROCEEDINGS{MAB_TaskSelection_Sima_2022,
  author={Sima, Qinghua and Gao, Guoju and Huang, He and Sun, Yu-E and Du, Yang and Wang, Xiaoyu and Wu, Jie},
  booktitle={Int. Conf. on Comp. Commun. and Netw. (ICCCN)}, 
  title={Multi-Armed Bandits Based Task Selection of A Mobile Crowdsensing Worker}, 
  year={2022},
  volume={},
  number={},
  pages={1-10},
  doi={10.1109/ICCCN54977.2022.9868929}}

@ARTICLE{Distributed_task_selection_Cheung_2021,
  author={Cheung, Man Hon and Hou, Fen and Huang, Jianwei and Southwell, Richard},
  journal={IEEE Trans. on Mob. Comput.}, 
  title={Distributed Time-Sensitive Task Selection in Mobile Crowdsensing}, 
  year={2021},
  volume={20},
  number={6},
  pages={2172-2185},
  doi={10.1109/TMC.2020.2975569}}

@ARTICLE{Task_allocation_Time_Constraint_XinLi_2021,
  author={Li, Xin and Zhang, Xinglin},
  journal={IEEE Trans. on Mob. Comput.}, 
  title={Multi-Task Allocation Under Time Constraints in Mobile Crowdsensing}, 
  year={2021},
  volume={20},
  number={4},
  pages={1494-1510},
  doi={10.1109/TMC.2019.2962457}}

@ARTICLE{Sparse_MCS_MARL_Chunyu_2024,
  author={Tu, Chunyu and Yu, Zhiyong and Han, Lei and Guo, Xianwei and Huang, Fangwan and Guo, Wenzhong and Wang, Leye},
  journal={IEEE Trans. on Mob. Comput.}, 
  title={Adaptive Budgeting for Collaborative Multi-Task Data Collection in Online Sparse Crowdsensing}, 
  year={2024},
  volume={23},
  number={7},
  pages={7983-7998},
  doi={10.1109/TMC.2023.3342206}}

@ARTICLE{Mobility_prediction_MCS_fuzzy_Zhang_2023,
  author={Zhang, Jinyi and Zhang, Xinglin},
  journal={IEEE Trans. on Mob. Comput.}, 
  title={Multi-Task Allocation in Mobile Crowd Sensing With Mobility Prediction}, 
  year={2023},
  volume={22},
  number={2},
  pages={1081-1094},
  doi={10.1109/TMC.2021.3088291}}

@ARTICLE{UAV_assisted_MCS_Gao_2023,
  author={Gao, Hui and Feng, Jianhao and Xiao, Yu and Zhang, Bo and Wang, Wendong},
  journal={IEEE Trans. on Mob. Comput.}, 
  title={{A UAV-Assisted Multi-Task Allocation Method for Mobile Crowd Sensing}}, 
  year={2023},
  volume={22},
  number={7},
  pages={3790-3804},
  doi={10.1109/TMC.2022.3147871}}

@article{stiglitz1977monopoly,
  title={Monopoly, non-linear pricing and imperfect information: the insurance market},
  author={Stiglitz, Joseph E},
  journal={The Review of Economic Studies},
  volume={44},
  number={3},
  pages={407--430},
  year={1977},
  publisher={Wiley-Blackwell}
}

@ARTICLE{Bernd_WSaad_OSL_2024,
  author={Simon, Bernd and Ortiz, Andrea and Saad, Walid and Klein, Anja},
  journal={IEEE Trans. on Commun.}, 
  title={Decentralized Online Learning in Task Assignment Games for Mobile Crowdsensing}, 
  year={2024},
  volume={72},
  number={8},
  pages={4945-4960},
  doi={10.1109/TCOMM.2024.3381718}
}

@INPROCEEDINGS{TMahn2021,
  author={Mahn, Tobias and Klein, Anja},
  booktitle={IEEE 10th International Conference on Cloud Networking (CloudNet)}, 
  title={A Global Orchestration Matching Framework for Energy-Efficient Multi-Access Edge Computing}, 
  year={2021},
  volume={},
  number={},
  pages={11-18},
  doi={10.1109/CloudNet53349.2021.9657120}}

@inproceedings{PPO_convergence_problem_2002,
author = {Kakade, Sham and Langford, John},
title = {Approximately Optimal Approximate Reinforcement Learning},
year = {2002},
isbn = {1558608737},
address = {San Francisco, CA, USA},
booktitle = {Proc. of the 19th Intl. Conf. on Machine Learning},
pages = {267–274},
numpages = {8},
}

@inproceedings{PPO_Convergence_and_optimality_Liu_2019,
author = {Liu, Boyi and Cai, Qi and Yang, Zhuoran and Wang, Zhaoran},
booktitle = {Advances in Neural Information Processing Systems},
 pages = {},
title = {Neural Trust Region/Proximal Policy Optimization Attains Globally Optimal Policy},
volume = {32},
year = {2019}
}

@ARTICLE{Simon2025Bargaining,
  author={Simon, Bernd and Adrian, Paul and Weber, Patrick and Felka, Patrick and Hinz, Oliver and Klein, Anja},
  journal={IEEE Trans. on Mob. Comput.}, 
  title={A Bargaining Approach for Service Placement in Multi-Access Edge Computing With Information Asymmetries}, 
  year={2025},
  volume={24},
  number={6},
  pages={5464-5481},
}

@misc{mcmahan2023communicationefficientlearningdeepnetworks,
      title={Communication-Efficient Learning of Deep Networks from Decentralized Data}, 
      author={H. Brendan McMahan and Eider Moore and Daniel Ramage and Seth Hampson and Blaise Agüera y Arcas},
      year={2023},
      eprint={1602.05629},
      archivePrefix={arXiv},
      primaryClass={cs.LG},
}

@misc{IPPO_deWitt_2020,
title={Is Independent Learning All You Need in the StarCraft Multi-Agent Challenge?}, 
author={Christian Schroeder de Witt and Tarun Gupta and Denys Makoviichuk and Viktor Makoviychuk and Philip H. S. Torr and Mingfei Sun and Shimon Whiteson},
year={2020},
eprint={2011.09533},
archivePrefix={arXiv},
primaryClass={cs.AI},
}

@article{yu2022surprising,
  title={{The surprising effectiveness of PPO in cooperative multi-agent games}},
  author={Yu, Chao and Velu, Akash and Vinitsky, Eugene and Gao, Jiaxuan and Wang, Yu and Bayen, Alexandre and Wu, Yi},
  journal={Advances in neural information processing systems},
  volume={35},
  pages={24611--24624},
  year={2022}
}

@ARTICLE{houda_FL_security_2024,
  author={Houda, Zakaria Abou El and Naboulsi, Diala and Kaddoum, Georges},
  journal={IEEE Internet of Things Journal}, 
  title={A Privacy-Preserving Collaborative Jamming Attacks Detection Framework Using Federated Learning}, 
  year={2024},
  volume={11},
  number={7},
  pages={12153-12164},
  doi={10.1109/JIOT.2023.3333870}}

@ARTICLE{Moudoud_FL_security_2024,
  author={Moudoud, Hajar and Houda, Zakaria Abou El and Brik, Bouziane},
  journal={IEEE Transactions on Consumer Electronics}, 
  title={Advancing Security and Trust in WSNs: A Federated Multi-Agent Deep Reinforcement Learning Approach}, 
  year={2024},
  volume={70},
  number={4},
  pages={6909-6918},
  doi={10.1109/TCE.2024.3440178}}

@ARTICLE{Nie_FLSecurity_2025,
  author={Nie, Xuefang and Wang, Chen and Zhou, Tianqing and Zhou, Qiangqiang and Zhu, Xusheng and Zhang, Jiliang},
  journal={IEEE Internet of Things Journal}, 
  title={Mobility-Aware Cooperative Caching in IoVs Based on Secure Asynchronous Federated and Deep Reinforcement Learning}, 
  year={2025},
  volume={12},
  number={12},
  pages={20572-20588},
  doi={10.1109/JIOT.2025.3544368}}
\end{spacing}

\end{document}